\documentclass[letterpaper, 10 pt, journal, twoside]{IEEEtran}
\usepackage[svgnames,table]{xcolor}
\usepackage{pifont}
\newcommand{\lilbox}[1]{\textcolor{#1}{\ding{110}}}
\newcommand{\xmark}{\ding{55}}%

\usepackage[caption=false]{subfig}
\usepackage{multirow}
\usepackage{booktabs}

\newcommand{\vect}[1]{\mathbf{#1}}

\ifCLASSINFOpdf
\usepackage{graphicx}
\else
\fi
\begin{document}
%
\title{Navigation-Oriented Scene Understanding for Robotic Autonomy: Learning to Segment Driveability in Egocentric Images}
%
%
%

\author{Galadrielle Humblot-Renaux$^{1}$, Letizia Marchegiani$^{2}$, Thomas B. Moeslund$^{1}$, and Rikke Gade$^{1}$%
\thanks{Manuscript received: September 8, 2021; Revised December 5, 2021; Accepted January 3, 2022.}
\thanks{This paper was recommended for publication by Editor Cesar Cadena Lerma upon evaluation of the Associate Editor and Reviewers' comments.
} 
\thanks{$^{1}$Galadrielle Humblot-Renaux, Thomas B. Moeslund and Rikke Gade are with the Visual Analysis and Perception Laboratory, Aalborg University, Denmark. {\tt\footnotesize \{gegeh,tbm,rg\}@create.aau.dk}}%
\thanks{$^{2}$Letizia Marchegiani is with the Department of Electronic Systems, Aalborg University, Denmark. {\tt\footnotesize {lm@es.aau.dk}}}%
\thanks{Digital Object Identifier (DOI): see top of this page.}
}
%
%

\markboth{IEEE Robotics and Automation Letters. Preprint Version. Accepted January, 2022}
{Humblot-Renaux \MakeLowercase{\textit{et al.}}: Navigation-Oriented Scene Understanding for Robotic Autonomy} 

%



\maketitle

\begin{abstract}
This work tackles scene understanding for outdoor robotic navigation, solely relying on images captured by an on-board camera. Conventional visual scene understanding interprets the environment based on specific descriptive categories. However, such a representation is not directly interpretable for decision-making and constrains robot operation to a specific domain. Thus, we propose to segment egocentric images directly in terms of how a robot can navigate in them, and tailor the learning problem to an autonomous navigation task. Building around an image segmentation network, we present a generic affordance consisting of 3 driveability levels which can broadly apply to both urban and off-road scenes. By encoding these levels with soft ordinal labels, we incorporate inter-class distances during learning which improves segmentation compared to standard ``hard'' one-hot labelling. In addition, we propose a navigation-oriented pixel-wise loss weighting method which assigns higher importance to safety-critical areas.
We evaluate our approach on large-scale public image segmentation datasets ranging from sunny city streets to snowy forest trails. In a cross-dataset generalization experiment, we show that our affordance learning scheme can be applied across a diverse mix of datasets and improves driveability estimation in unseen environments compared to general-purpose, single-dataset segmentation.
\end{abstract}


%
\IEEEpeerreviewmaketitle

\section{Introduction}
%
%
%
%
\IEEEPARstart{A}{robot} roaming outdoors ``in the wild'' needs to know where to go, and what to avoid. It may traverse vast areas with unfamiliar terrain, unexpected obstacles or challenging environmental conditions which degrade its view, yet should still be able to identify safe and suitable terrain to drive on. In this work, our aim is to parse images captured by an outdoor robot and interpret them at the pixel level in order to inform navigation decisions~\cite{vision-actions-2019}, without constraining scene understanding to a specific domain. We rather consider an ``open-world'' navigation task spanning on-road and off-road scenes, from grassy fields to city traffic, or from forest trails to pedestrian areas. In this context, it is beneficial to know not only where the road/path is (if there is one), but also what other areas are driveable, although perhaps not ideally so.

\begin{figure}[!t]
    \centering
    \includegraphics[width=\linewidth]{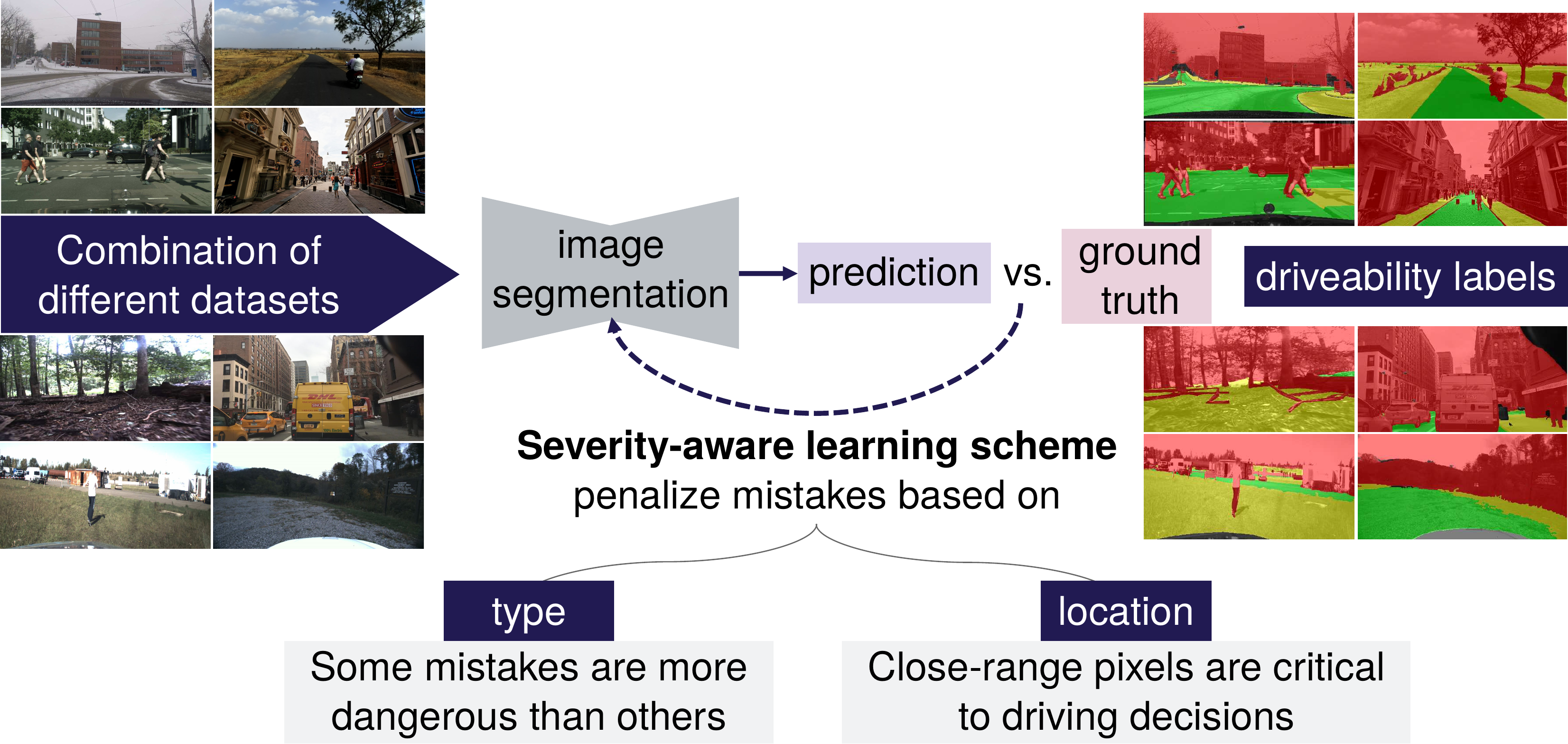}
    \caption{Overview of our navigation-oriented learning scheme for learning 3-level driveability across diverse outdoor scenes from pixel-annotated datasets.}
    \label{fig:method-overview}
\end{figure}

To learn a consistent and useful representation across diverse scenes, we interpret images directly in terms of potential action rather than object descriptions, following the concept of visual \textit{affordance}~\cite{visual-affordance-survey-2021}. Existing affordance learning approaches for outdoor navigation rely on sensor data recorded on a real platform to generate self-supervised or weakly-supervised labels~\cite{path-proposals-segmentation-2017,legged-terrain-classification-2019,badgr-2021} - this is an impractical and resource-intensive process, which limits the diversity of images seen during training. In contrast, we approach affordance learning as a fully supervised image segmentation problem, leveraging the abundance of large-scale scene understanding datasets. We present a 3-level driveability affordance which is directly interpretable for robotic decision-making and applies across arbitrary outdoor environments (not just roads as in~\cite{path-proposals-segmentation-2017}, or static off-road scenes as in~\cite{badgr-2021,legged-terrain-classification-2019}), while explicitly tailoring the learning problem to navigation. Our contributions include:
\begin{enumerate}
    \item a navigation-oriented framework which enables cross-dataset training, bypassing the need for real-world exploration or additional labelling effort;
    \item a soft label encoding which incorporates the ambiguity and order between levels of driveability,  penalizing some mis-classifications more than others during learning;
    \item a loss weighting scheme which, rather than treating all pixels as equally important for navigation, concentrates learning in safety-critical areas while allowing leniency around object outlines and distant scene background;
    \item a challenging experimental procedure: beyond same-dataset testing, we evaluate the generalizability of our approach on three unseen datasets, including the WildDash benchmark~\cite{dataset-wilddash-2018} which captures a large variety of difficult driving scenarios across 100+ countries.
\end{enumerate}
Figure~\ref{fig:method-overview} illustrates the core idea of our approach. This learning scheme is, to the best of our knowledge, the first attempt at incorporating an inter-class ranking in a scene understanding task, taking both the \textit{type} and \textit{location} of mistakes into account during learning to improve affordance segmentation.

%

\section{Related work}\label{sec:related}

Semantic segmentation for outdoor scene understanding is extensively studied~\cite{cv-autonomous-driving-2020}. The bulk of existing approaches either segment images at the object level~\cite{segnet-2017}, or reduce the problem to binary segmentation (e.g. road vs. rest~\cite{multinet-2018} or free space vs. obstacles \cite{stixelnet-2015}). Object-based approaches are dataset- and domain-specific, unnecessarily descriptive for navigation, hinder generalisation~\cite{visual-abstraction-driving-2020}, and scale poorly to unseen obstacles or unstructured scenes~\cite{mseg-2020}. Conversely, binary segmentation is much more generic, but does not capture the kinds of degrees of driveability which are relevant for off-road robotic applications traversing diverse terrain~\cite{badgr-2021,legged-terrain-classification-2019,fuzzy-traversability-2020}.

Instead, we take a probabilistic affordance segmentation approach to scene understanding. Existing works in this direction are either contained to simulation~\cite{vizdoom-affordance-maps-2020}, indoor environments~\cite{multiscale-affordance-seg-2016,timo-affordance-seg-2019} or static outdoor scenes~\cite{legged-terrain-classification-2019}. In contrast, our approach considers challenging, dynamic outdoor scenes captured by a real robot or vehicle. Closely related to our work,~\cite{path-proposals-segmentation-2017} proposes to segment obstacles and a proposed path in driving scenes, with weakly-supervised labels generated from Lidar and odometry data, and unlabeled pixels assigned to a 3rd ``unknown'' class. While~\cite{path-proposals-segmentation-2017} achieves remarkable performance in structured urban scenes, the driveable area is limited to the current lane, and the method's applicability to off-road scenarios is unclear. Like~\cite{path-proposals-segmentation-2017}, we adopt a 3-class definition for scene understanding, but as recommended by~\cite{affordance-dev-survey-2016}, we introduce a \textit{degree} of driveability. This allows us to generate viable predictions beyond on-road driving scenes, with the aim of enabling open-world robot navigation.

More generally, our method contrasts with existing affordance learning approaches for outdoor navigation which require additional sensor data to be collected by a navigation platform (eg. Lidar~\cite{stixelnet-2015,path-proposals-segmentation-2017,badgr-2021}, odometry~\cite{path-proposals-segmentation-2017,legged-terrain-classification-2019}, IMU~\cite{badgr-2021}, or force-torque measurements~\cite{legged-terrain-classification-2019}). Instead, our method leverages the wide range of readily available image segmentation datasets at no annotation cost, and only requires monocular images at training-time. 

In addition, as illustrated in Figure~\ref{fig:method-overview}, our approach places particular emphasis on generalization and mistake severity for safe robotic perception. This contrasts with all the aforementioned works, which are limited to single-dataset training/evaluation and treat all pixels and classes as interchangeable during learning. For bridging the gap across different segmentation datasets, rather than expanding the label space to accommodate an ever-increasing number of object labels as in~\cite{mseg-2020}, we reduce the label space down to 3 generic driveability levels. Our ``severity-aware'' segmentation framework builds upon the findings in~\cite{wasserstein-severity-aware-2020}, which show that encoding the severity of different misclassifications in ground truth labels significantly reduces the risk of collision. However, we also consider the \textit{location} of mistakes during learning, and propose a multi-domain affordance-based representation which is tailored to robotic navigation.


\section{Approach}

Our approach primarily revolves around how we formulate the learning problem. First, we generate driveability labels by mapping object-based pixel annotations from existing semantic segmentation datasets to a 3-level affordance. ``Hard'' driveability labels are then softened to model inter-class severity. Lastly, our loss weighting scheme selectively emphasizes the areas most relevant to navigation during learning.

\subsection{From object semantics to driveability labels}\label{sec:driv-levels}

We define a 3-level affordance to characterize the driveability~\cite{driveability-assessment-2020} of a pixel:
\begin{itemize}
    \item \lilbox{green} \textit{Preferable}: where we expect the robot to drive (paved roads or paths);
    \item \lilbox{yellow} \textit{Possible}, but not preferable: areas which are technically navigable but more challenging or less suitable, and would not be chosen as a first resort (e.g. grass, sand);
    \item \lilbox{red} \textit{Impossible} or undesirable: any part of the scene which is unreachable (e.g. the sky) or should be unconditionally avoided (obstacles, hazardous terrain).
\end{itemize}

\begin{figure}[b]
   \centering
   \includegraphics[width=\linewidth]{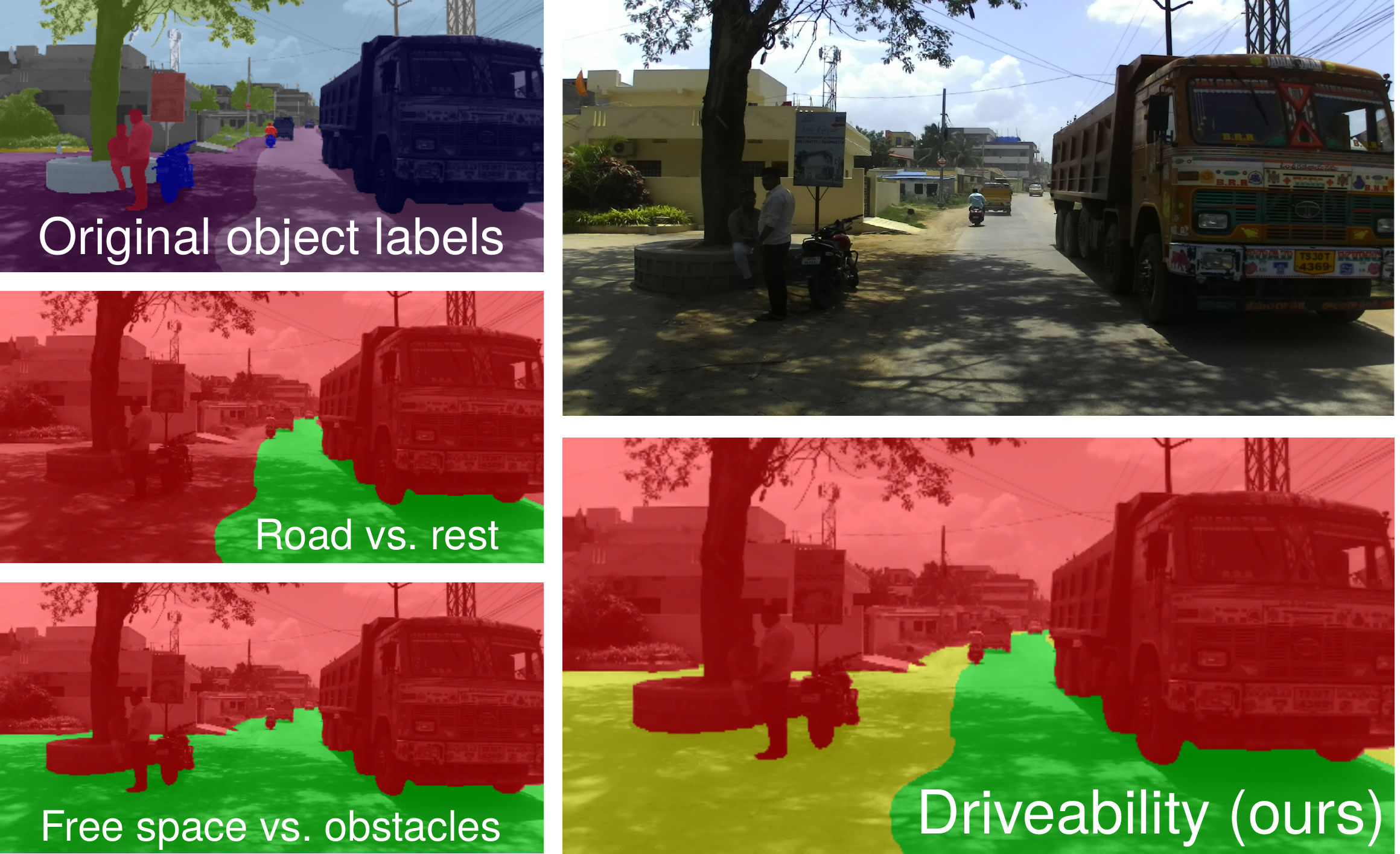}
    
    \caption{Example of a pixel-annotated outdoor scene from the IDD dataset~\cite{dataset-idd-2019}. We map the original object classes to driveability levels.}
    \label{fig:labels-idd}
\end{figure}

\begin{figure}[b]
    \centering
    \subfloat[One-hot labels]{%
    \includegraphics[width=0.4\linewidth]{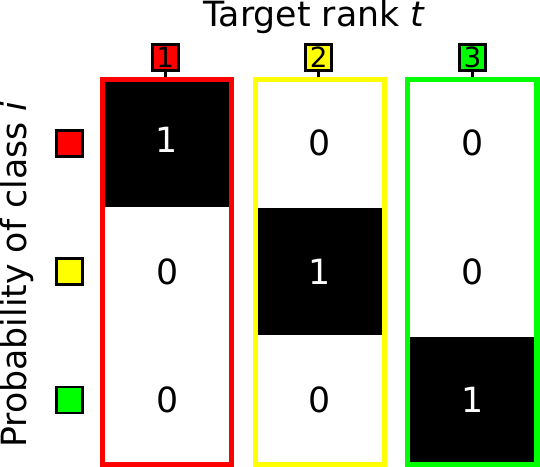}
    \label{fig:onehot}
    }
    \hspace{1em}
    \subfloat[SORD labels]{%
    \includegraphics[width=0.4\linewidth]{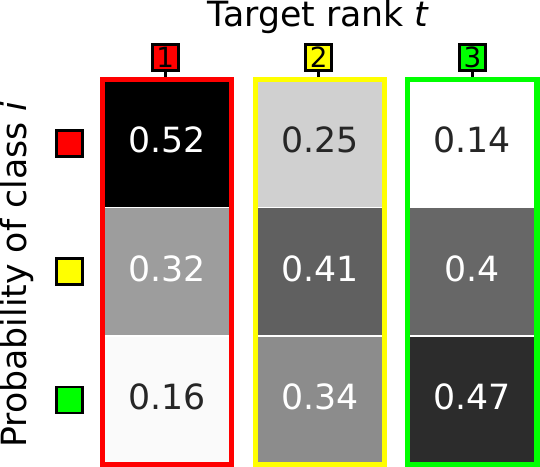}
    \label{fig:sord}
    }
    \caption{Label class probabilities with a standard one-hot encoding (left) vs. the SORD labelling scheme (right).}
    \label{fig:sord-labels}
\end{figure}

This taxonomy is inspired by the action plausibility ratings proposed in~\cite{action-plausibility-2020}. For each pixel, we generate an affordance label by mapping its original semantic label (eg. car, sidewalk, tree, road) to a driveability level \{\lilbox{red}, \lilbox{yellow}, \lilbox{green}\}. As discussed in \cite{affordances-rob-survey-2020}, such a mapping from descriptive semantic labels to affordance is somewhat reductive as it does not take any contextual information into account - however, it remains a common approach~\cite{multiscale-affordance-seg-2016,timo-affordance-seg-2019}, since it enables fully-supervised learning without the need for manual affordance labelling.

In our experiments, we compare this 3-level definition to two common binary segmentation approaches mentioned in Section~\ref{sec:related}: road vs. rest segmentation (\lilbox{yellow} level mapped to \lilbox{red}) and free space vs. obstacles segmentation (\lilbox{yellow} mapped to \lilbox{green}). A comparison is shown in Figure~\ref{fig:labels-idd}. The intermediate level \lilbox{yellow} can serve as a fallback in the absence of a clear path in the scene, and leaves more flexibility at the planning level (e.g. if a robot has an off-road navigation target, or if an autonomous vehicle needs to park, change lanes, or overturn a car).

\subsection{From one-hot to soft ordinal labels}

Intuitively, mis-classifying an area which is \lilbox{green} \textit{preferable} (e.g. the path) to drive on as \lilbox{red} \textit{impossible} should be penalized more heavily than classifying it as \lilbox{yellow} \textit{possible}. However, a standard one-hot encoding (Figure~\ref{fig:onehot}) coupled with a categorical loss function do not capture this distinction during learning: mis-classifications are treated as equally severe regardless of the target. To incorporate a notion of pair-wise distance or severity between driveability levels, we opt for a soft labelling approach, which does not require any architectural changes and has shown to improve generalization in a wide range of tasks~\cite{does-label-smoothing-help-2020}. Specifically, we implement the Soft Ordinal vectors (or SORD) labelling scheme proposed in~\cite{sord-2019}: standard one-hot encoded labels are converted to a softmax-normalized probability distribution based on a ranking definition, such that the target class has the highest probability and the other probabilities encode a distance from the target class. Given a set of ranks $R = \{r_{impossible}, r_{possible}, r_{preferable}\}$ (one per driveability level), a SORD ground truth label $\vect{y}$ is generated based on a target rank $r_t$ as follows:

\begin{equation}\label{eq:sord}
y_i = \frac{\exp \left( -\phi(r_t,r_i)\right)}{\sum_{k \in R} \exp \left(-\phi(r_t,r_k)\right)} \quad \forall r_i \in R
\end{equation}

\noindent where $\phi(r_t,r_i)$ is a metric function which penalizes deviation from the target rank $r_t$. As inter-rank distances approach infinity, $\vect{\hat{y}}$ reduces to a one-hot encoded vector; as the distances approach 0, $\vect{\hat{y}}$ approaches a uniform probability distribution.

For this application, we consider a simple ranking definition between driveability levels: $R = \{$\lilbox{red}~$1,~$\lilbox{yellow}$~2,~$\lilbox{green}$~ 3\}$ (least to most driveable). We define the metric penalty function $\phi$ as the square log difference (SLD) $\phi(r_t,r_i) = |\log_e(r_i) - \log_e(r_t)|^2$ , which reduces the penalty with increasing rank.  Compared to absolute difference for instance, SLD shifts the middle rank \lilbox{yellow} \textit{possible} much ``closer'' to \lilbox{green} \textit{preferable} than to \lilbox{red} \textit{impossible}: in other words, the distinction between obstacles and driveable areas is much more clear-cut than the blurry line between driveable areas which are \lilbox{green} \textit{preferable} or not. Intuitively, this mirrors the ambiguity that a human annotator would face when labelling images: we are much less likely to hesitate when categorizing an area as obstacle vs. non-obstacle than when determining whether a driveable area is preferable or not.

Figure~\ref{fig:sord-labels} shows the resulting asymmetrical SORD label encoding $\vect{y}$ for each of the 3 possible driveability targets, compared to a one-hot categorical encoding. Following \cite{sord-2019}, we then take the loss per pixel as the Kullback-Leibler (KL) divergence between the predicted class probability vector $\vect{\hat{y}}$ and the SORD label $\vect{y}$ from~(\ref{eq:sord}): 
$L_{KL}(\vect{y}||\vect{\hat{y}}) = \sum_{\forall r_i \in R} y_i \log_e \frac{y_i}{\hat{y}_i}$

\begin{figure}[!t]
    \centering
    \includegraphics[width=0.7\linewidth]{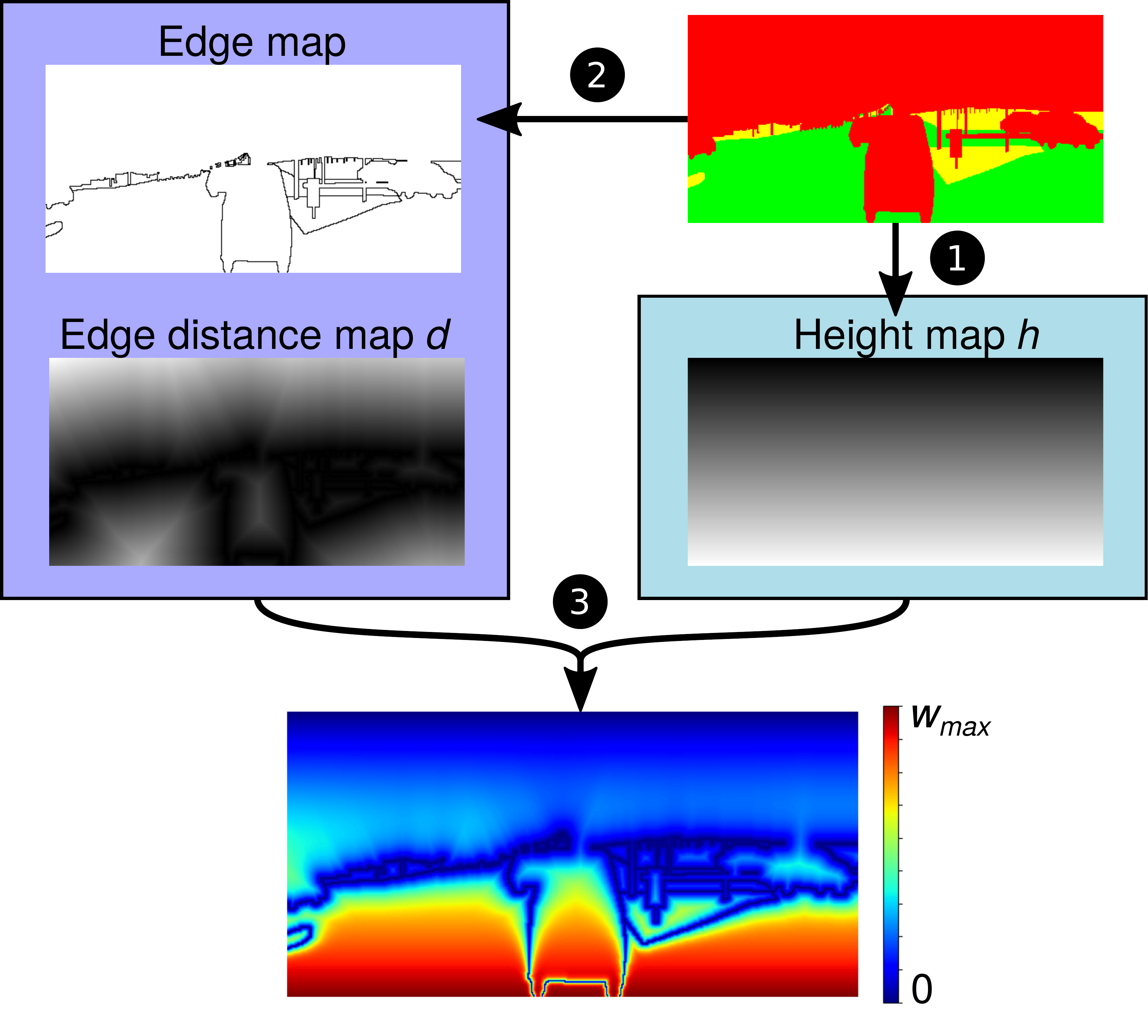}
    \caption{Steps in the loss weight map computation, numbered and illustrated with a ground truth sample from the Kitti dataset~\cite{kitti-2013}.}
    \label{fig:lw-kitti}
\end{figure}

\subsection{Loss weighting}\label{sec:lw}
We argue that for navigation, detailed understanding of the entire scene is not necessary. Rather than giving each pixel equal contribution, we focus learning \textit{away} from object contours which are difficult to learn~\cite{boundary-relaxation-2019}, and towards areas in the camera's immediate vicinity which are critical to driving decisions. For selectively emphasizing relevant pixels during learning, we adapt the loss weighting scheme proposed in~\cite{unet-2015}. We adapt its formulation to our task such that boundary pixels are assigned the lowest weight, and we introduce a notion of image depth to distinguish between close-range and background elements. Given a pixel location $\vect{p} = [p_x,p_y]^T$ in the ground truth mask, we compute a weight map $w(\vect{p})$ which is applied to the loss per pixel via element-wise multiplication. The weight of a pixel depends on its Euclidean distance $d(\vect{p})$ to the closest segmentation boundary and on its vertical position (height) in the image $h(\vect{p})$:

\begin{equation}\label{eq:loss-weight-map}
    w(\vect{p}) = h(\vect{p}) \cdot \left[ 1 - \exp \left(- \frac{d(\vect{p})}{1 + \beta (1 - h(\vect{p})^2)^2} \right) \right]
\end{equation}

\noindent where $\beta$ is an experimentally defined constant. The height map $h(\vect{p})$ is used to scale the rate at which the pixel weight increases when moving away from a boundary, and as a pixel-wise multiplication factor which assigns higher weight to lower pixels. It serves as a naive placeholder for depth data, under the simple assumption that the lower a pixel in the image, the closer it is to the camera.

As illustrated in Figure~\ref{fig:lw-kitti}, we generate a weight map $w(\vect{p})$ from a ground truth mask in three steps:
\begin{enumerate}
    \item the height map $h(\vect{p})$ is pre-computed for all possible pixel locations based on the image height H as: $h(\vect{p}) = p_y / H$ such that pixels in the lowest row of the image have the value 1 and top row pixels have a value of 0.
    \item for computing the edge distance map $d(\vect{p})$, we first perform edge detection on the gray-scaled ground truth mask, binarize the edge map, and apply a distance transform~\cite{distance-transform-cv-1986} with a $ 5\times5 $ kernel.
    \item the weight map $w(\vect{p})$ is then computed following~(\ref{eq:loss-weight-map}), and min-max normalized to lie within $[0,w_{max}]$.
\end{enumerate}

\section{Experimental set-up}

\subsection{Architecture and hyper-parameters} For pixel-wise classification, we pick SegNet~\cite{segnet-2017} as a base network,  similarly to~\cite{path-proposals-segmentation-2017}. Our variant applies drop-out (rate of 0.5) in the six deepest encoder and decoder blocks for regularization, and reduces the number of convolutional layers in each block to 2 (as opposed to 3 in the deepest blocks of VGG-16~\cite{vgg-2015}), resulting in a total of 20 convolutional layers. We measure a forward pass time of 32ms on the NVIDIA TITAN X for a single sample. In all our experiments, we train SegNet using Adam optimization~\cite{adam-2015} ($\beta_1 = 0.9$, $\beta_2 = 0.999$) to minimize the KL divergence. Unlabeled/void pixels are ignored: the batch loss is computed as the sum of the loss per non-void pixel, divided by the number of non-void pixels in the batch. Samples are fed to the network in shuffled mini-batches of size 8, and the best model is selected based on minimal validation loss.

\subsection{Cross-domain datasets}

\begin{table}[b]
\centering
   \aboverulesep=0ex
   \belowrulesep=0ex
\caption{Cross-domain combination of image segmentation datasets used in our zero-shot cross-dataset experiment.}
\label{tab:datasets}
\resizebox{\linewidth}{!}{%
\begin{tabular}{c|c|c}
\toprule
\rule{0pt}{1.1EM}%
\textit{scene type} & \textbf{Training \& validation} \hfill (\# images) & \textbf{Testing} \hfill (\# images) \\
 \midrule
 \rule{0pt}{1.1EM}%
\multirow{4}{*}{urban driving} &
Cityscapes~\cite{cityscapes-2016}\hfill (3,484)
 &
Kitti~\cite{kitti-2013} \hfill (200) \\
& BDD~\cite{dataset-bdd-2020}\hfill(8,000) & \\
& Mapillary~\cite{dataset-mapillary-2017} \hfill (20,000) &\\
& ACDC~\cite{dataset-acdc-2021}\hfill(2,006) &
\\
\midrule
\rule{0pt}{1.1EM}%

\multirow{3}{*}{\begin{tabular}[c]{@{}c@{}}unstructured / \\ off-road\end{tabular}}
& RUGD~\cite{dataset-rugd-2019} \hfill (5492) & Freiburg Forest~\cite{multispectral-scene-understanding-forest-2017} \hfill (366)\\
& YCOR~\cite{dataset-ycor-2018} \hfill (1076) & \\
& TAS500~\cite{dataset-tas500-2021} \hfill (540) & \\

\midrule
\rule{0pt}{1.1EM}
mixed & IDD~\cite{dataset-idd-2019} \hfill (8089) &
WildDash~\cite{dataset-wilddash-2018} \hfill (4256)\\
\bottomrule
\end{tabular}
}
\end{table}

Our approach is entirely data-driven: accurate estimates of driveability in unconstrained environments require challenging samples to be included during training. For evaluating generalization to new environments with our method, we adopt a similar zero-shot cross-dataset strategy to~\cite{mseg-2020}: models are trained on a combination of cross-domain datasets, and evaluated on a separate combination of datasets which have never been seen during training or validation.

We select outdoor scene understanding datasets with pixel-level annotations and RGB images captured by a vehicle or mobile robot, as outlined in Table~\ref{tab:datasets}. For training, we include Cityscapes~\cite{cityscapes-2016}, a widespread benchmark featuring ``clean'' scenes, as well as more recent driving datasets covering a wide range of environmental conditions, sensor characteristics and geographical contexts including Mapillary~\cite{dataset-mapillary-2017}, Berkeley DeepDrive (BDD)~\cite{dataset-bdd-2020} and ACDC~\cite{dataset-acdc-2021}. Outside of urban landscapes, RUGD~\cite{dataset-rugd-2019}, YCOR~\cite{dataset-ycor-2018}  and TAS500~\cite{dataset-tas500-2021} cover off-road grassy environments. Lastly, IDD~\cite{dataset-idd-2019} brings a unique challenge since it was captured in unstructured Indian traffic and rural scenes. For evaluation, we select 3 datasets with varying levels of difficulty. Kitti~\cite{kitti-2013} is a small-scale benchmark of ``clean'' city driving scenes. Freiburg Forest~\cite{multispectral-scene-understanding-forest-2017} was captured by a mobile robot traversing forested trails, with some challenging illumination conditions, but no dynamic obstacles. WildDash~\cite{dataset-wilddash-2018} was specifically designed as a difficult test set for evaluating robustness to visual driving hazards in diverse environments. We use each dataset's official train/validation split during learning, and full datasets during testing, resulting in a total of 42,759 / 5,939 / 4,822 images for training/validation/testing respectively. 

Note that these 11 datasets were annotated under different sets of semantic classes, but mapping their original object labels to a generic notion of driveability allows us to bridge this semantic gap during training and evaluation. During learning, each driveability level is informed by samples from all 8 training datasets. To counteract the imbalance in dataset size, similarly to~\cite{mseg-2020}, mini-batches are constructed with an equal number of samples (1 in our case) from each dataset.

\subsection{Data preparation}

\textbf{Input color:} While it is commonplace to preserve color information in input images for scene understanding~\cite{adaptnet-2017,cv-autonomous-driving-2020}, we speculate that color may add unnecessary or distracting information when trying to learn such an abstract concept as driveability. Thus, we investigate the importance of color in our experiments by comparing the standard RGB representation with a single-channel grayscale input.

\textbf{Input size:} This is also an important consideration, with a trade-off between computational cost and segmentation detail. Resizing images to fixed dimensions is common practice, especially when learning from a combination of datasets~\cite{mseg-2020}. For our affordance learning task, retaining a high level of detail is not a primary concern, but incorporating global context is crucial~\cite{timo-affordance-seg-2019}. Therefore, we opt for a small input resolution of $240 \times 480$ - the same width as in~\cite{segnet-2017}, but with a wider aspect ratio to accommodate wide-FOV datasets.

\textbf{Data augmentation:} During training, input samples are randomly augmented on-the-fly with geometric (horizontal flip, rotation, crop, perspective transform, grid-based distortion) and photometric (brightness, contrast, tone curve and color manipulation) transformations, each having a probability of $0.5$. See~\cite{albumentations-2020} for a detailed description.

\subsection{Training procedure}\label{sec:training}

\textbf{Pre-training}: As a starting point, we train SegNet on Cityscapes to predict the 30 original object classes in the dataset~\cite{cityscapes-2016}, using an initial learning rate of $10^{-3}$. We refer to this model as Cityscapes\textit{\textsubscript{obj}}. Note that this model is trained under a standard learning scheme (one-hot labels, uniformly weighted loss), and thus can be substituted with other pre-trained segmentation models.

\textbf{Driveability via transfer learning}: To learn 3-level driveability from a combination of datasets, the last convolutional layer of Cityscapes\textit{\textsubscript{obj}} is re-initialized with 3 output channels and trained under the SORD labelling scheme with an initial learning rate of $10^{-4}$ until convergence.
    
\textbf{Loss weighting (LW)} is implemented as a final training stage to consolidate the segmentation, while maintaining the same labelling scheme. Weight maps are generated with $w_{max} = 10$ (same as~\cite{unet-2015}) and $\beta = 30$.

\subsection{Evaluation metrics}

\textbf{Classification metrics:} In the context of autonomous navigation, under-segmentation of obstacles and over-segmentation of driveable areas pose particular safety risks. Therefore, aligning with~\cite{path-proposals-segmentation-2017}, we select two segmentation metrics of interest: pixel-wise recall (R) for the \lilbox{red} level, and precision (P) for \lilbox{green}. We also introduce a \textit{weighted} version of these metrics R\textit{\textsubscript{w}} and P\textit{\textsubscript{w}} which weighs each pixel  based on the LW map, thus emphasizing the most navigation-relevant areas.

\textbf{Regression metrics:} The segmentation metrics above do not capture inter-rank distances. Therefore, similarly to~\cite{sord-2019}, we report Root-mean-square error (RMSE) to evaluate the degree of error between predicted and ground truth driveability levels, with heavier penalty for large error (confusion between \lilbox{green} and \lilbox{red} levels). Based on~\cite{better-mistakes-2020}, we also compute a measure of \textit{mistake severity} (MS) as the mean absolute error of \textit{incorrect} predictions; note that MS is fully decoupled from accuracy. We normalize MS per pixel, such that it ranges from 0 to 1: mis-classifying a pixel as \lilbox{yellow} yields a MS of 0, while confusing the \lilbox{red} and \lilbox{green} levels yields a MS of 1.

\begin{figure*}[!hbt]
   \centering
    \includegraphics[width=\textwidth]{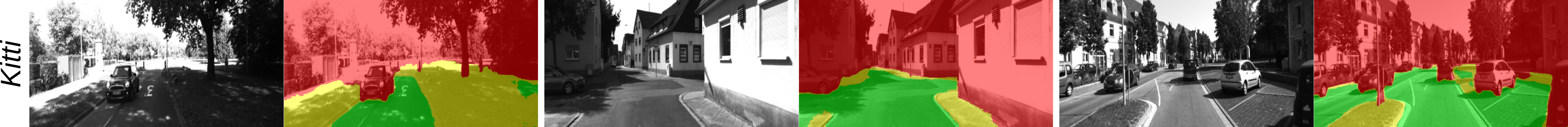}
    
    \vspace{0.5em}
    
    \includegraphics[width=\textwidth]{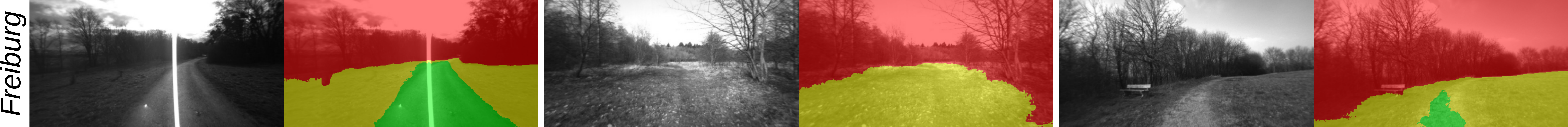}
    
    \vspace{0.5em}
    
    \includegraphics[width=\textwidth]{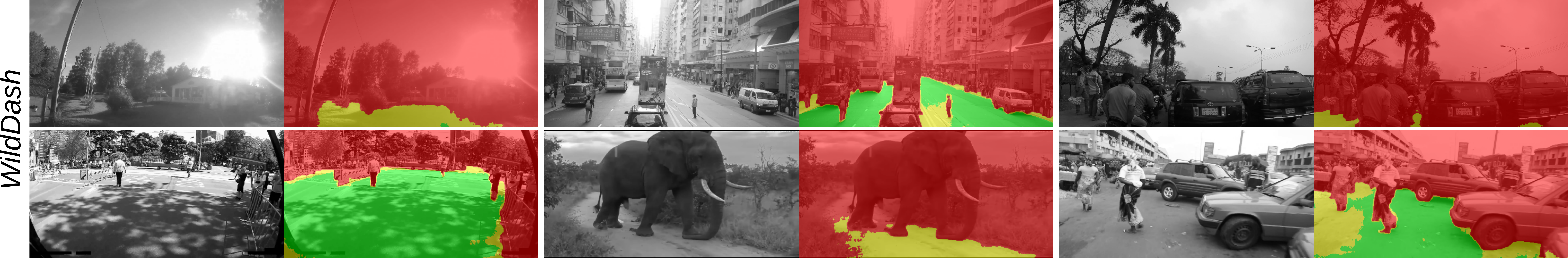}
    \caption{Examples of out-of-dataset predictions by the proposed model, trained on the cross-domain dataset with soft driveability labels and loss weighting.}
    \label{fig:examples-lw-ood}
\end{figure*}

\section{Results}

We first validate our 3-level driveability definition and learning scheme. We then benchmark our approach against the state-of-the-art and an object-based baseline, and comment on the effect of input color to motivate the use of grayscale images in our experiments. Lastly, we show some failure cases and probabilistic predictions of our model.

\begin{figure}[!b]
    \centering
    
    \includegraphics[width=\linewidth]{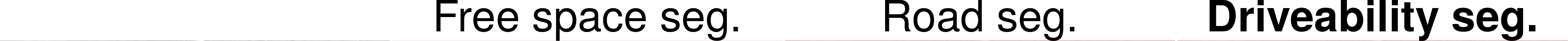}
    \includegraphics[width=\linewidth]{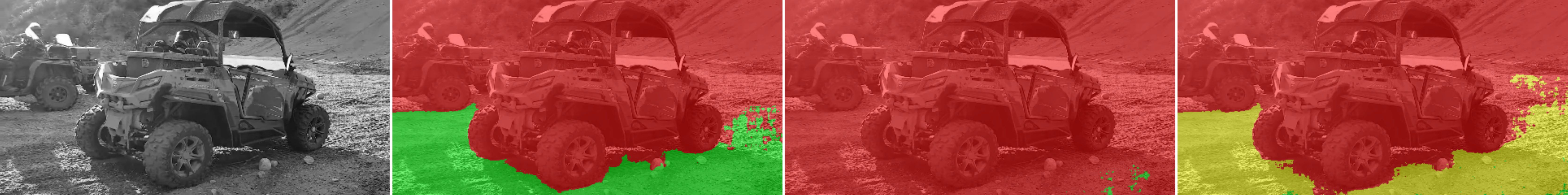}
    \includegraphics[width=\linewidth]{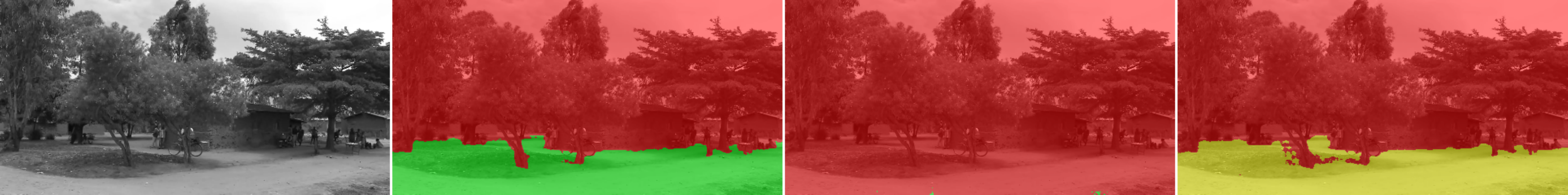}
    \small{\textit{Road segmentation fails to produce viable predictions in open areas}}
    
    \vspace{0.5em}
    
    \includegraphics[width=\linewidth]{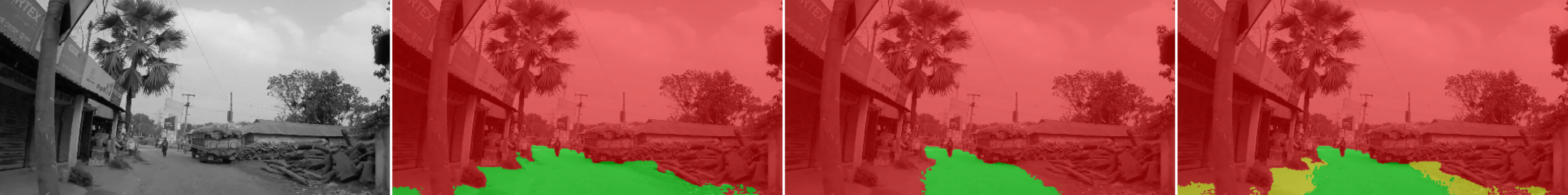}
    \includegraphics[width=\linewidth]{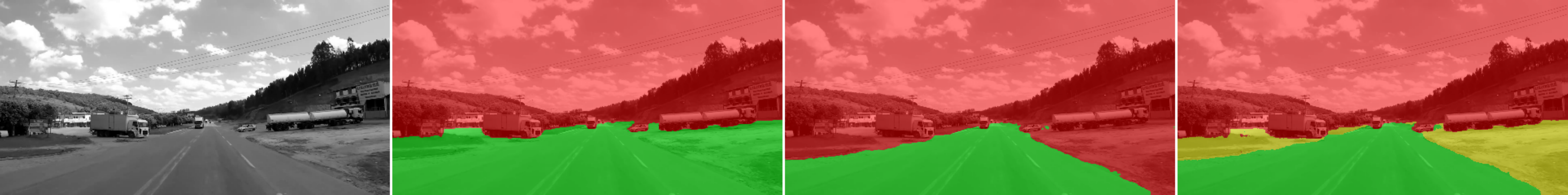}
    \small{\textit{Free space segmentation treats all ground as equally driveable}}
    
    \caption{Out-of-dataset predictions by one-hot cross-domain models on WildDash samples, trained under different class definitions.}
    \label{fig:binary-comp}
\end{figure}

\textbf{3-level driveability vs. binary segmentation:} For comparison, we train a cross-domain model with standard one-hot labels scheme for each of the three class definitions presented in Figure~\ref{fig:labels-idd}. Table~\ref{tab:binary-comp} compares the models' performance in terms of segmentation error (calculated with ranks $R = \{$~\lilbox{red}~$1,$~\lilbox{green}~$3\}$ for the binary segmentation baselines). The driveability model consistently achieves the lowest segmentation error, followed by free space segmentation in urban scenes and road/path segmentation in mixed or forested scenes. Figure~\ref{fig:binary-comp} shows the qualitative advantage of our 3-level driveability definition over these binary segmentation approaches. The driveability model identifies \lilbox{yellow} \textit{possible} driveable ground in off-road or open areas, while also distinguishing \lilbox{green} \textit{preferable} areas when there is a clear path in the scene.

\begin{table}[!htb]
\caption{RMSE of one-hot cross-domain driveability models compared to binary segmentation baselines.}
\label{tab:binary-comp}
\resizebox{\linewidth}{!}{%
\begin{tabular}{@{}lcccc@{}}
\toprule
Segmentation class definition & Cross-domain (val) & Kitti & Freiburg & WildDash \\ \midrule
Road/path \lilbox{green} vs. \lilbox{red} rest  & 0.412 & 0.423 & 0.310 & 0.490 \\
\addlinespace[0.1em]
Free space \lilbox{green} vs. \lilbox{red} obstacles  & 0.437 & 0.377 & 0.445 & 0.407 \\
\addlinespace[0.1em]
\textbf{3-level driveability} \lilbox{green} \lilbox{yellow} \lilbox{red} & \textbf{0.283} & \textbf{0.311} & \textbf{0.284} & \textbf{0.402} \\ \bottomrule
\end{tabular}%
}
\end{table}

\begin{figure}[!b]
\centering
\includegraphics[width=\linewidth]{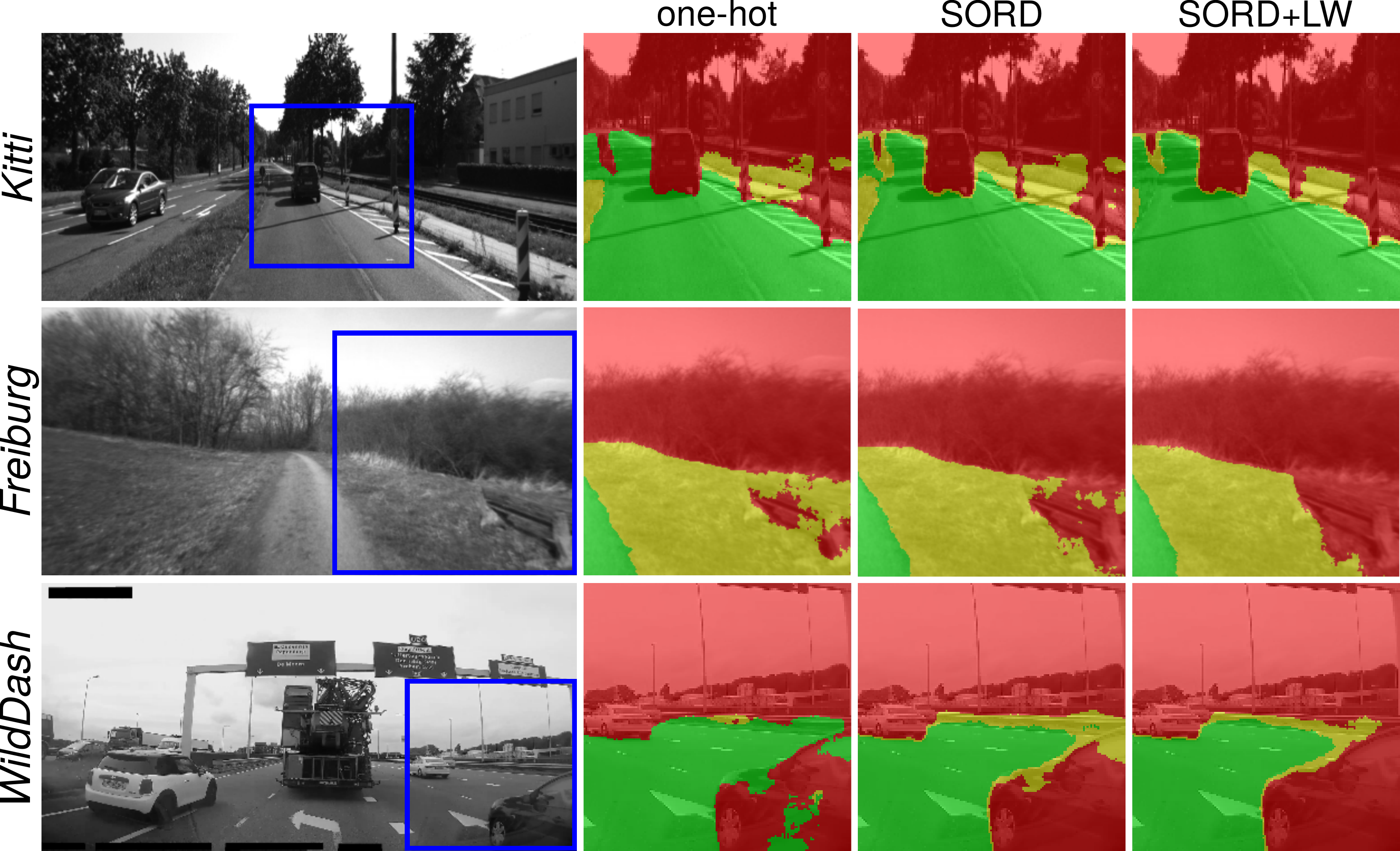}

    \caption{Selected crops of out-of-dataset predictions by the cross-domain driveability model, showing the qualitative effect of the soft labelling and loss weighting training schemes.}
    \label{fig:examplessord}
\end{figure}

\textbf{Navigation-oriented learning scheme:} Figure~\ref{fig:examples-lw-ood} shows predictions by our proposed cross-domain SORD+LW model on out-of-dataset samples from the three unseen test sets, and we include a video showing additional qualitative results as supplementary material. The model produces sensible driveability estimates across a wide range of navigation scenarios, with variations in scene layout and contents, lighting and weather conditions, as well as camera characteristics and perspectives. Table~\ref{tab:results-ood} reports quantitative performance, with a comparison between a model trained under our proposed training scheme (Section~\ref{sec:training}), and a standard model trained with one-hot labels in the transfer learning stage and uniformly-weighted pixel-wise loss. Table~\ref{tab:results-ood} shows our navigation-oriented learning scheme to be effective at bringing down RMSE on the validation set and on every unseen test dataset, with SORD labelling reducing mistake severity by almost 50\% compared to a standard one-hot model. The addition of LW consistently improves segmentation in safety-critical areas, as indicated by the weighted obstacle \lilbox{red} recall and \lilbox{green} precision scores. We note the most significant quantitative improvement in generalization performance for Freiburg Forest's highly unstructured environments, where our method helps disambiguate the fuzzy transitions between path, grass and surrounding vegetation without getting lost in details. Looking at the overall aspect of the segmentation across test samples, we find that SORD labelling produce smoother contours and less spotty segmentation, and encourages cautious, low-stakes predictions especially for ambiguous border pixels. As can be seen in the examples of Figure~\ref{fig:examplessord}, this visually manifests as a layer of \lilbox{yellow} pixels around non-driveable areas, rather than sharp transitions between \lilbox{red} and \lilbox{green} levels. While this deviates from what ground truth masks look like, we consider it beneficial for navigation, since it essentially adds a safe margin around obstacles. LW, which concentrates learning away from distant details towards close-range and non-border areas, results in a more approximate but cohesive segmentation. 

\begin{table}[!hbt]
\caption{Same-dataset and zero-shot generalization performance of our cross-domain driveability models.}
\label{tab:results-ood}

\resizebox{\linewidth}{!}{%
\begin{tabular}{@{}p{0.16\linewidth}p{0.18\linewidth}p{0.09\linewidth}p{0.11\linewidth}p{0.09\linewidth}>{\centering\arraybackslash}p{0.11\linewidth}p{0.09\linewidth}p{0.09\linewidth}@{}}
\toprule
\textit{Test data} & \textit{Learning} & \lilbox{red} R \% & \lilbox{red} R\textit{\textsubscript{w}} \% & \lilbox{green} P \% & \lilbox{green} P\textit{\textsubscript{w}} \% & MS \% & RMSE \\
 \midrule
 
 \multirow{3}{*}{\begin{tabular}[c]{@{}l@{}}Cross-domain\\(validation)\end{tabular}}
& one-hot       & \textbf{98.41} & 97.77        & 93.20         & 94.32         & 15.28         & 0.283 \\
& SORD          & 98.12         & 97.48         & 93.04         & 93.70         & \textbf{7.94} & \textbf{0.276} \\
& SORD + LW     & 98.33         & \textbf{97.88} & \textbf{93.75} & \textbf{94.71} & 9.15       & 0.278 \\ 

 \midrule
 
\multirow{3}{*}{\begin{tabular}[c]{@{}c@{}}Kitti\end{tabular}}
& one-hot       & 98.72         & 98.17             & 87.86             & 90.18             & 10.79         & 0.311 \\
& SORD          & 98.42         & 97.95             & \textbf{89.55}    & 90.82             & \textbf{5.79} & \textbf{0.293} \\
& SORD + LW     & \textbf{98.79} & \textbf{98.64}   & 89.44             & \textbf{91.09}    & 7.43          & 0.304 \\ 

 \midrule
 
\multirow{3}{*}{\begin{tabular}[c]{@{}c@{}}Freiburg\end{tabular}}
& one-hot       & 94.15         & 89.98             & 85.19             & 88.27             & 1.75          & 0.284 \\
& SORD          & 96.12         & 94.07            & 80.38             & 83.15            & \textbf{0.50} & 0.269 \\
& SORD + LW     & \textbf{97.57} & \textbf{96.98}   & \textbf{86.29}  & \textbf{89.26}      & 0.69          & \textbf{0.258} \\ 

 \midrule
\multirow{3}{*}{\begin{tabular}[c]{@{}c@{}}WildDash\end{tabular}}
& one-hot       & \textbf{98.71}& 98.07             & 91.63 & 93.68 & 30.27 & 0.402 \\
& SORD          & 98.18         & 97.46            & 93.95
& 95.25 & \textbf{15.64}  & \textbf{0.369} \\
& SORD + LW     & 98.58         & \textbf{98.08}    & \textbf{94.01} & \textbf{95.48} & 18.54 & 0.380 \\ \bottomrule
\end{tabular}%
}

\end{table}

\textbf{Comparison to the state-of-the-art:} The closest candidate for comparison are the segmentation results in~\cite{path-proposals-segmentation-2017} for the general \textit{obstacle} class, defined similarly to our \lilbox{red} \textit{impossible} level. The authors train a SegNet model on weakly labelled images from Kitti Raw~\cite{kitti-2013}, and evaluate it on the Kitti Object \& Tracking datasets (over 85k obstacle bounding boxes). We evaluate our cross-domain SORD+LW model with the same procedure and metrics in Table~\ref{tab:oxford-kitti-detection}. Note that the pixel recall metric considers the whole bounding box area, while instance recall requires a certain percentage of pixels in a box to be predicted as \lilbox{red} \textit{obstacle} for it to be considered detected. Our model achieves state-of-the-art object detection ($>50\%$ instance recall) on this dataset, despite not having seen any Kitti images during training. The lower pixel recall and $>75\%$ instance recall scores of our model can be attributed to the granularity of our labels and predictions compared to~\cite{path-proposals-segmentation-2017}'s coarse, ``boxy'' segmentations, which naturally take up a larger portion of the ground truth bounding boxes on this benchmark.

\begin{table}[!b]
\caption{Obstacle segmentation results on the Kitti Object \& Tracking datasets (evaluation procedure from~\cite{path-proposals-segmentation-2017}).}
\label{tab:oxford-kitti-detection}
\resizebox{\linewidth}{!}{%
\begin{tabular}{@{}cccccc@{}}
\toprule
 &
 \begin{tabular}[c]{@{}c@{}}\textit{Seen Kitti}\\\textit{images?}\end{tabular} & \textit{Input} &
 Pixel recall & \begin{tabular}[c]{@{}c@{}}Instance recall\\ (\textgreater 50\%)\end{tabular} & \begin{tabular}[c]{@{}c@{}}Instance recall\\ (\textgreater 75\%)\end{tabular} \\ \midrule
\cite{path-proposals-segmentation-2017} & 24,443 & RGB & \textbf{93.53\%} & 99.55\% & \textbf{97.93\%} \\
ours & \xmark & Gray & 88.09\% & \textbf{99.74\%} & 96.34\% \\ \bottomrule
\end{tabular}%
}
\end{table}

In terms of qualitative results, while~\cite{path-proposals-segmentation-2017} fails to predict viable path segmentations in ambiguous road configurations (e.g. tight turns in intersections) and does not show results in road-free scenes, the examples in Figures~\ref{fig:examples-lw-ood} and \ref{fig:binary-comp} show that our model successfully identifies  \textit{preferable} \lilbox{green} driveable areas even in the absence of a structured lane, while falling back to the \lilbox{yellow} level in open unstructured areas.

\begin{table}[!b]
\caption{RMSE of our model compared to single-dataset baselines.}
\label{tab:cross-vs-single-mse}
\resizebox{\linewidth}{!}{%
\begin{tabular}{@{}llcccc@{}}
\toprule
\multicolumn{1}{c}{\textit{Train data}} & \textit{Learning} & Cityscapes (val) & Kitti & Freiburg & WildDash \\ \midrule
Cityscapes\textit{\textsubscript{obj}} & one-hot & 0.210 & 0.353 & 0.660 & 0.491 \\
\addlinespace[0.1em]
Cityscapes\textit{\textsubscript{driv}} & SORD+LW & \textbf{0.207} & 0.317 & 0.491 & 0.402 \\
\addlinespace[0.1em]
Cross-domain\textit{\textsubscript{driv}} & SORD+LW & 0.226 & \textbf{0.304} & \textbf{0.258} & \textbf{0.380} \\ \bottomrule
\end{tabular}%
}
\end{table}

\textbf{Comparison to an object-based single-dataset baseline:} The conventional approach to semantic scene segmentation consist of learning object descriptions on a single dataset. In contrast, our driveability definition allows us to combine heterogeneously labelled datasets during training. To show the benefit of our approach for generalization to new scenes, we take Cityscapes\textit{\textsubscript{obj}} as an experimental baseline, and map its object-based predictions to driveability levels for evaluation. We then apply the transfer learning and LW stages to learn driveability on Cityscapes (Cityscapes\textit{\textsubscript{driv}}). Table~\ref{tab:cross-vs-single-mse} compares our cross-domain model with these two single-dataset baselines. Comparing the two Cityscapes models, we note that learning driveability with SORD+LW consistently reduces same-dataset and generalization error compared to a one-hot object-based approach, with the most notable improvement for Cityscapes $\rightarrow$ Freiburg Forest transfer. Extending the findings in~\cite{mseg-2020}, our results show cross-domain learning to be beneficial for segmenting driveability in out-of-dataset images: learning driveability across a diverse 8-dataset combination reduces generalization error across all 3 unseen test datasets. While the performance of Cityscapes models drops when faced with Freiburg Forest's unstructured scenes, the cross-domain models maintain an RMSE below 0.4 (and pixel accuracy above 90\%) across all test sets.

\begin{figure}[!htb]
   \centering
    \includegraphics[height=0.14\textheight]{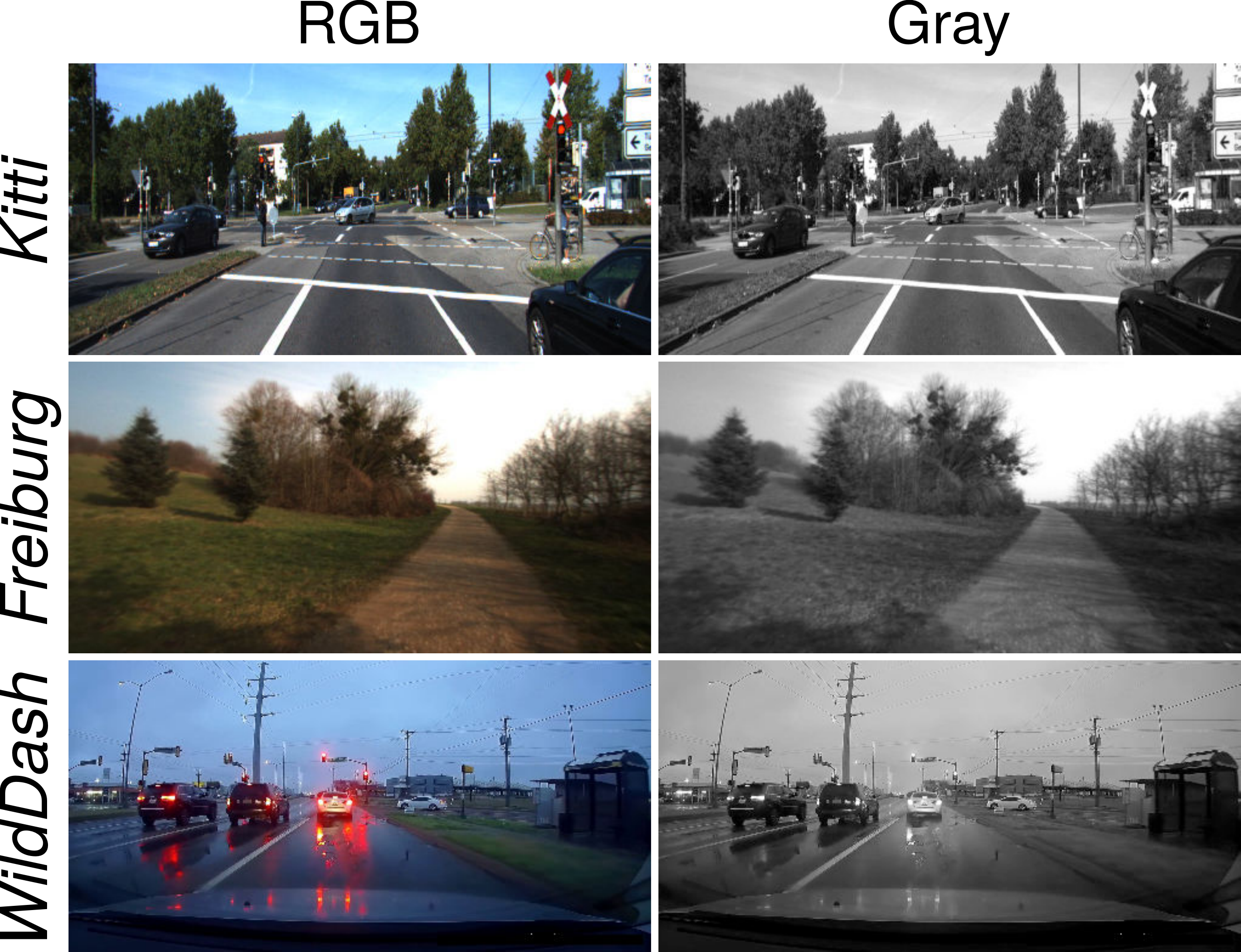}
    \includegraphics[height=0.14\textheight]{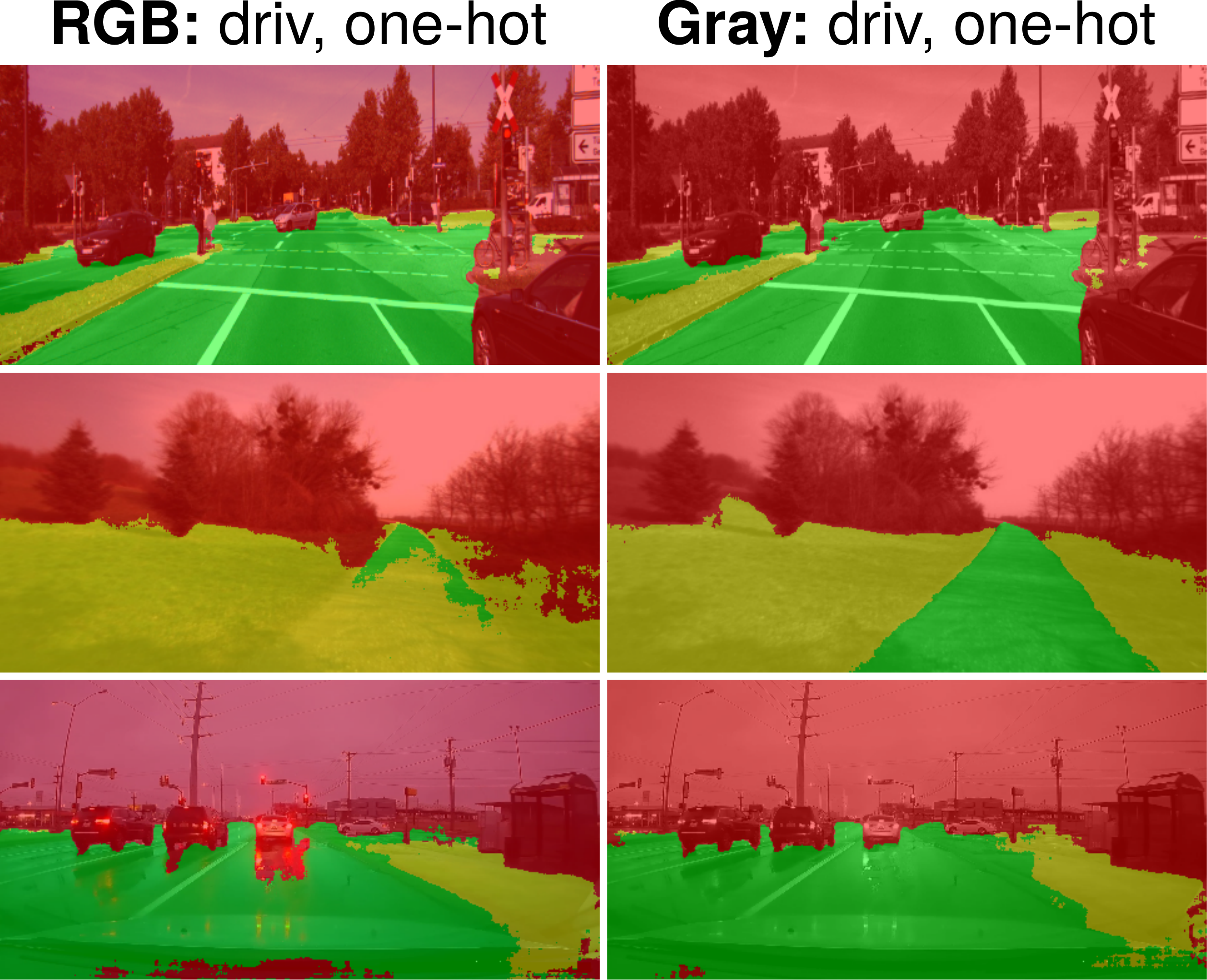}
    \caption{Qualitative comparison of out-of-dataset predictions by the cross-domain model trained on RGB vs. grayscale input images.}
    \label{fig:examples-color-vs-gray}
\end{figure}

\textbf{Does color matter?} On unseen samples from a known dataset or from a dataset captured in ideal urban scenarios (Cityscapes and Kitti in Table~\ref{tab:color} and Figure~\ref{fig:examples-color-vs-gray}), color brings a small improvement in segmentation. However, interestingly, we note a significant performance gap in favour of grayscale models when evaluating zero-shot generalization to challenging new scenes (Freiburg Forest and WildDash). While grayscale models are blind to dataset-specific color palettes, RGB models seem to make color-class associations (e.g. dark gray for the driveable road, bright red for cars) which may not hold in different outdoor environments (e.g. brown paths in Freiburg Forest, red reflections on the road).

\begin{table}[!htb]
\caption{Effect of input color on \lilbox{red} recall for one-hot models.}
\label{tab:color}
\centering
\resizebox{0.9\linewidth}{!}{%
\begin{tabular}{@{}llcccc@{}}
\toprule
\multicolumn{1}{c}{\textit{Train data}} & \textit{Input} & Cityscapes (val) & Kitti & Freiburg & WildDash \\ \midrule
\multirow{2}{*}{Cityscapes\textit{\textsubscript{obj}}} & RGB & \textbf{99.51} & \textbf{99.30} & 89.11 & 92.62 \\
 & Gray & 98.91 & 97.96 & \textbf{92.53} & \textbf{93.36} \\ \midrule
\multirow{2}{*}{Cross-domain\textit{\textsubscript{driv}}} & RGB & \textbf{99.33} & \textbf{98.91} & 91.55 & 92.36 \\
 & Gray & 99.32 & 98.72 & \textbf{94.11} & \textbf{98.71} \\ \bottomrule
\end{tabular}%
}
\end{table}

\begin{figure}[!t]
    \centering
    
    \includegraphics[width=0.246\linewidth]{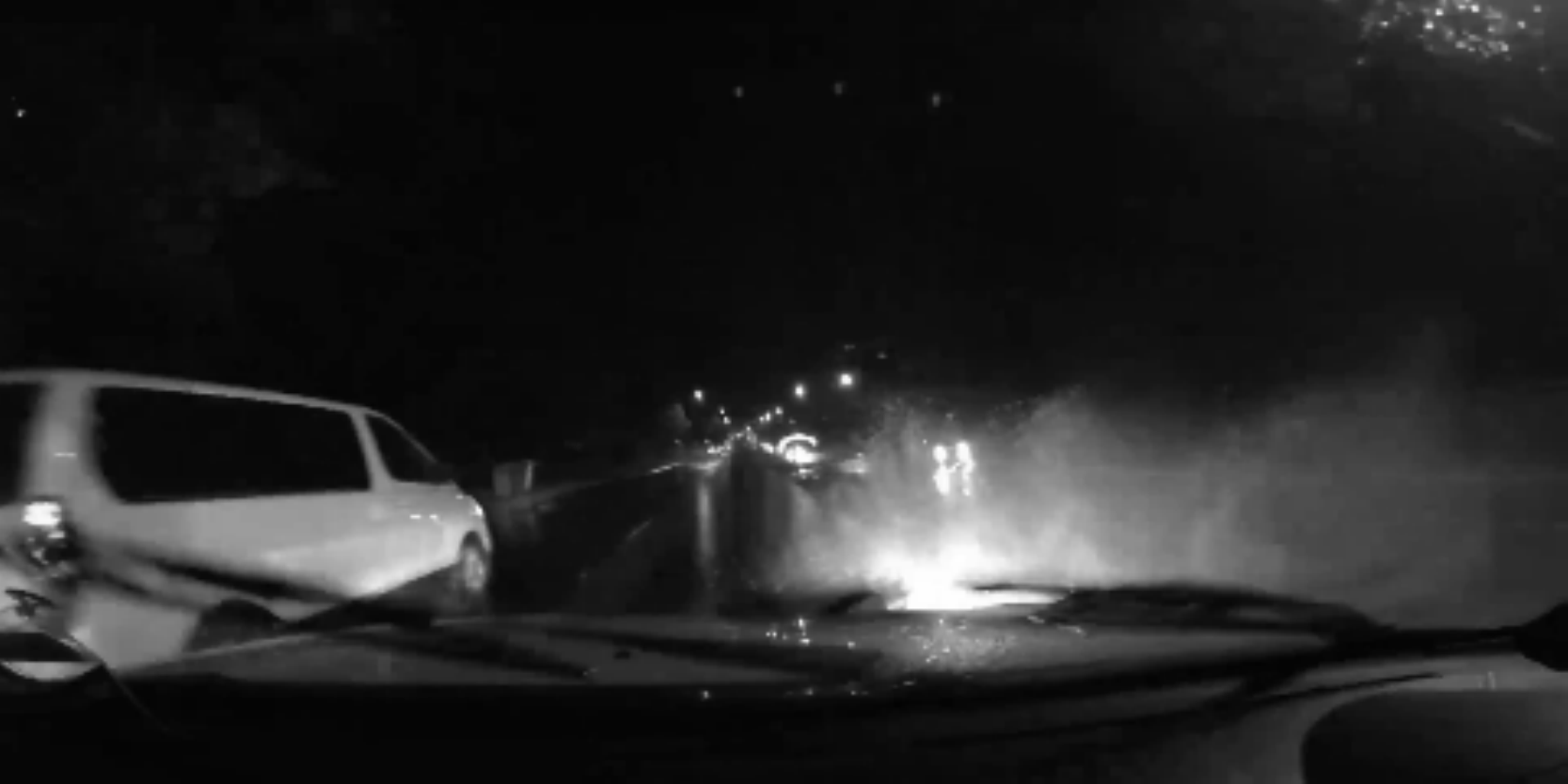}\includegraphics[width=0.246\linewidth]{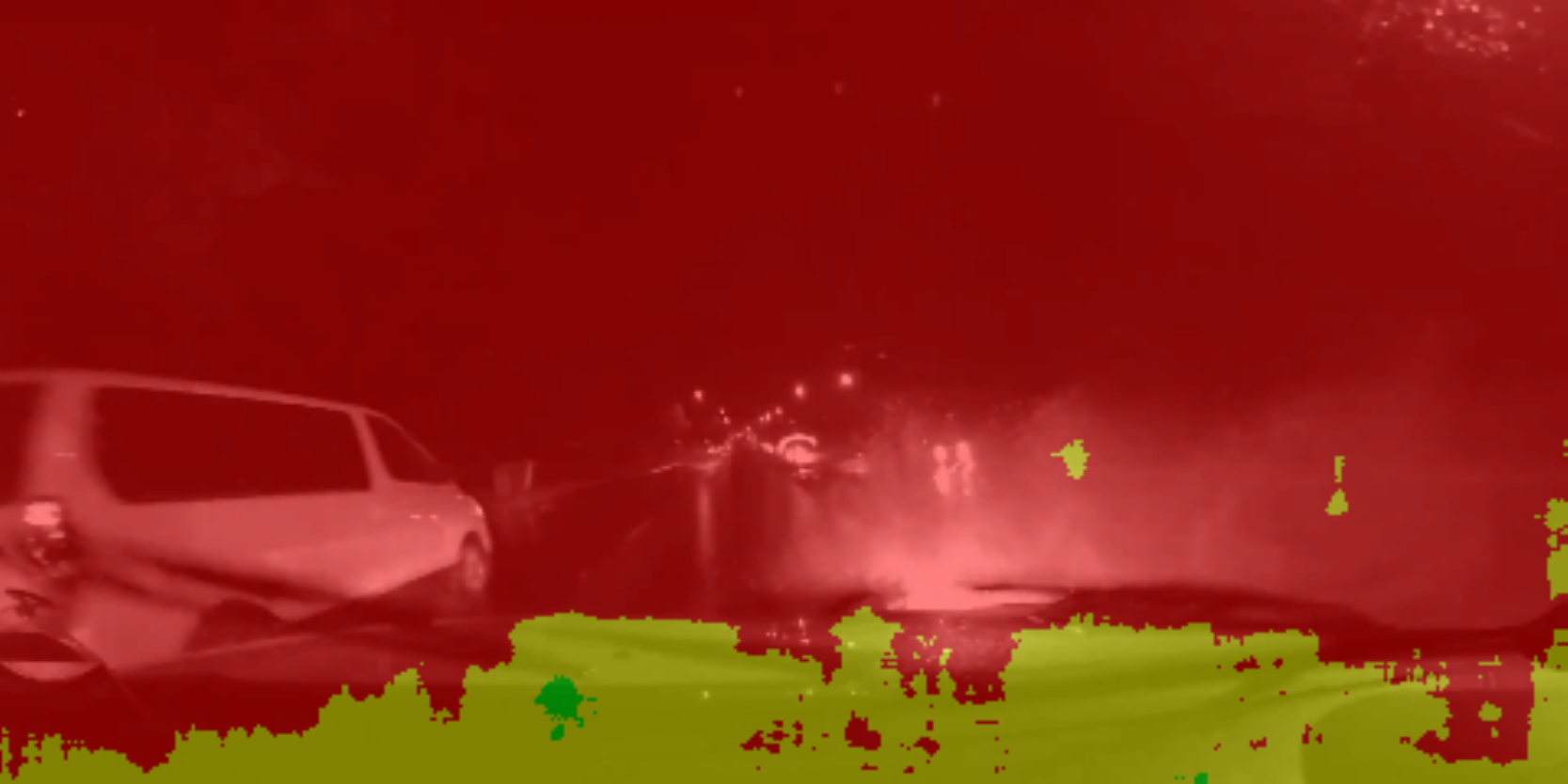}
    \includegraphics[width=0.246\linewidth]{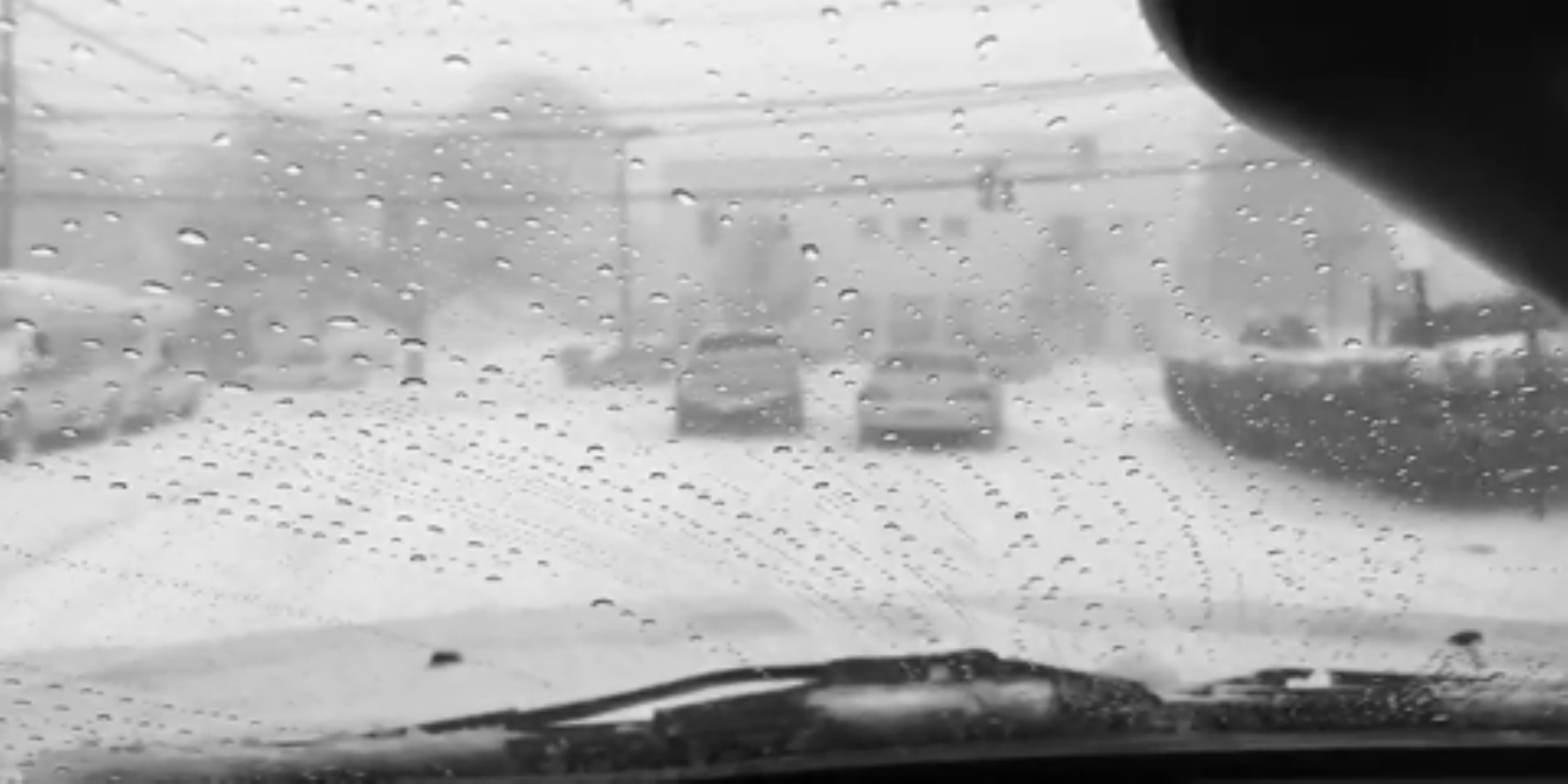}\includegraphics[width=0.246\linewidth]{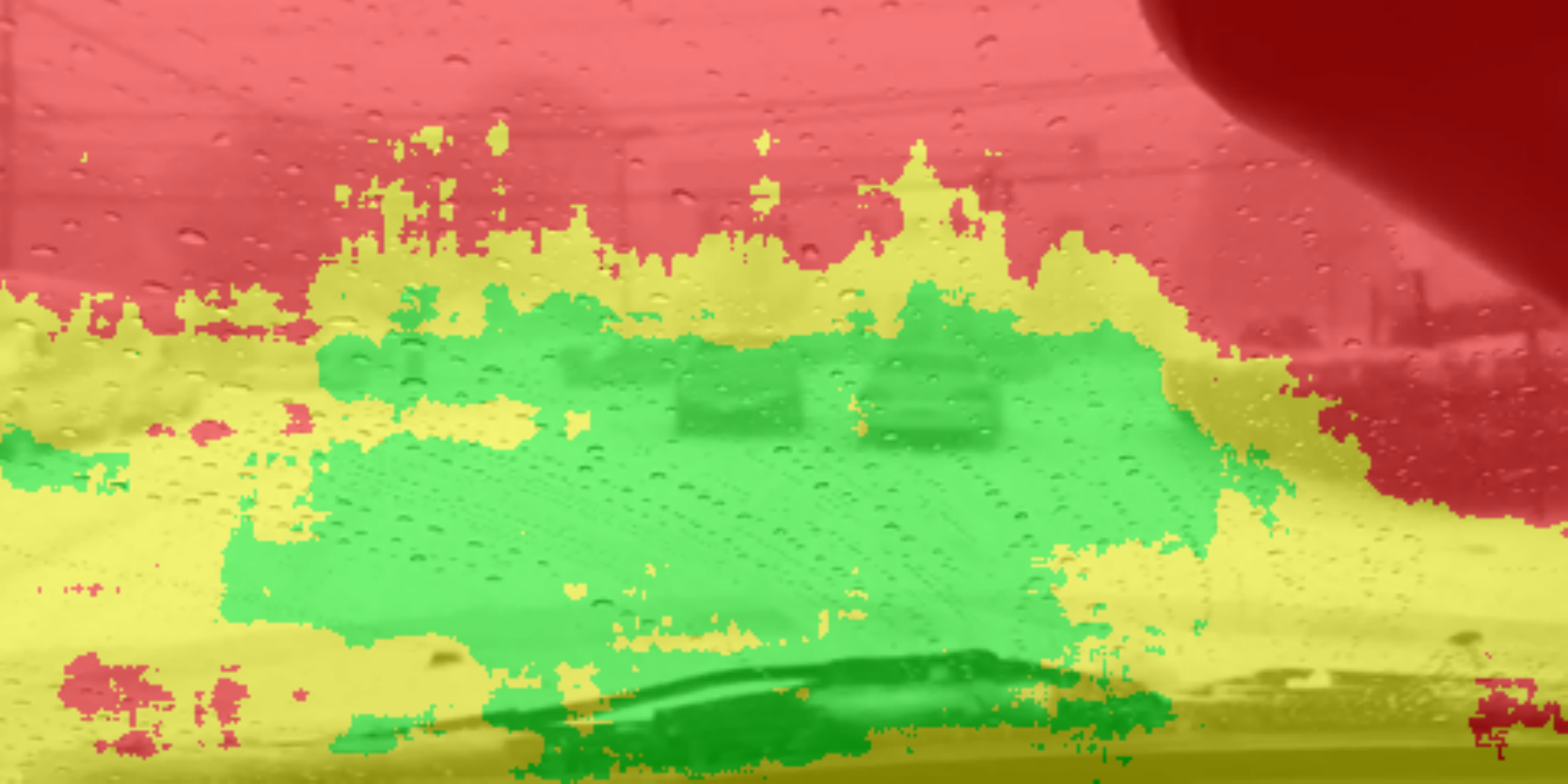}
    \small{\textit{Blinded by the weather}}
    
    \vspace{0.5em}
    
    \includegraphics[width=0.246\linewidth]{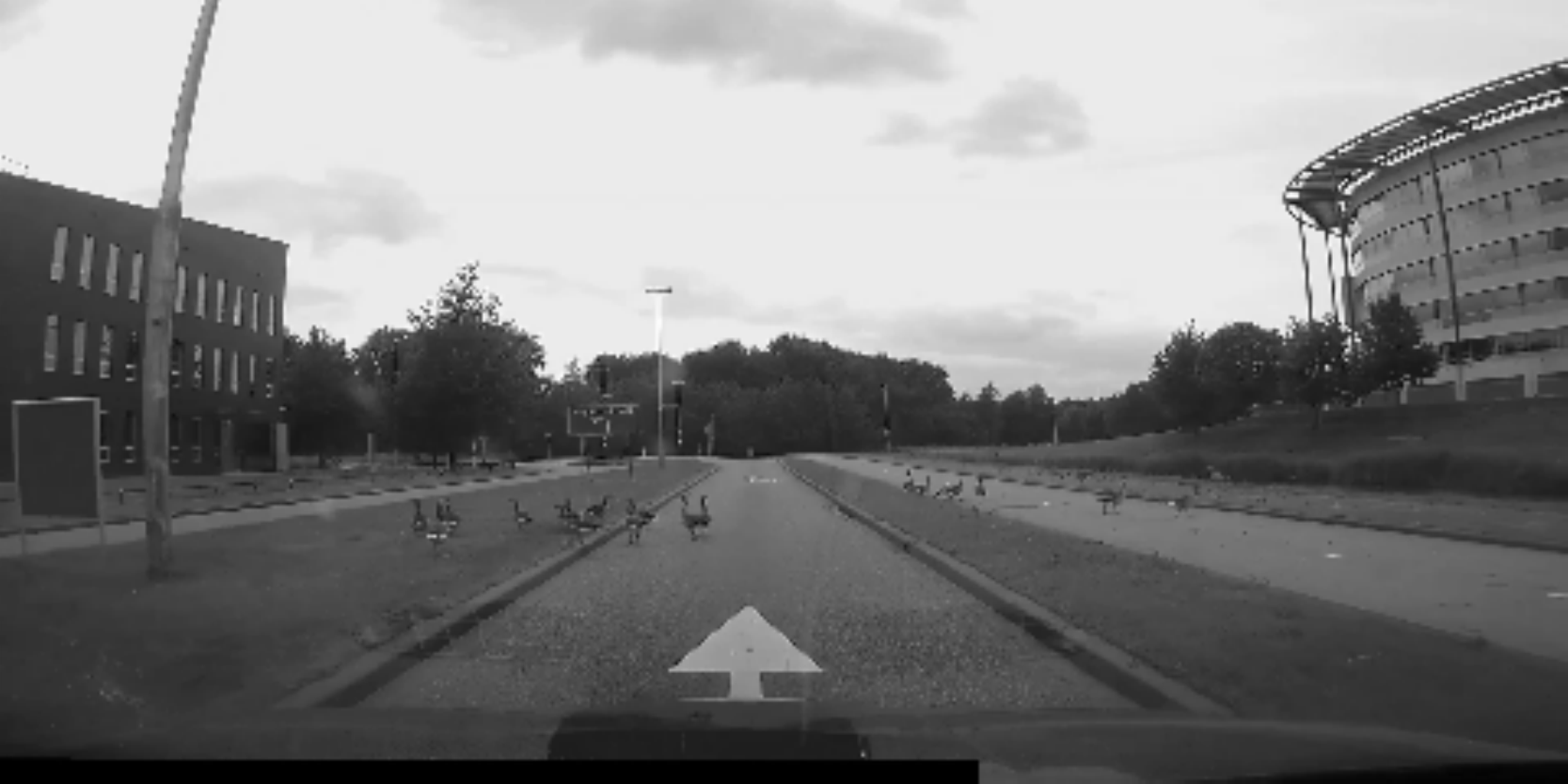}\includegraphics[width=0.246\linewidth]{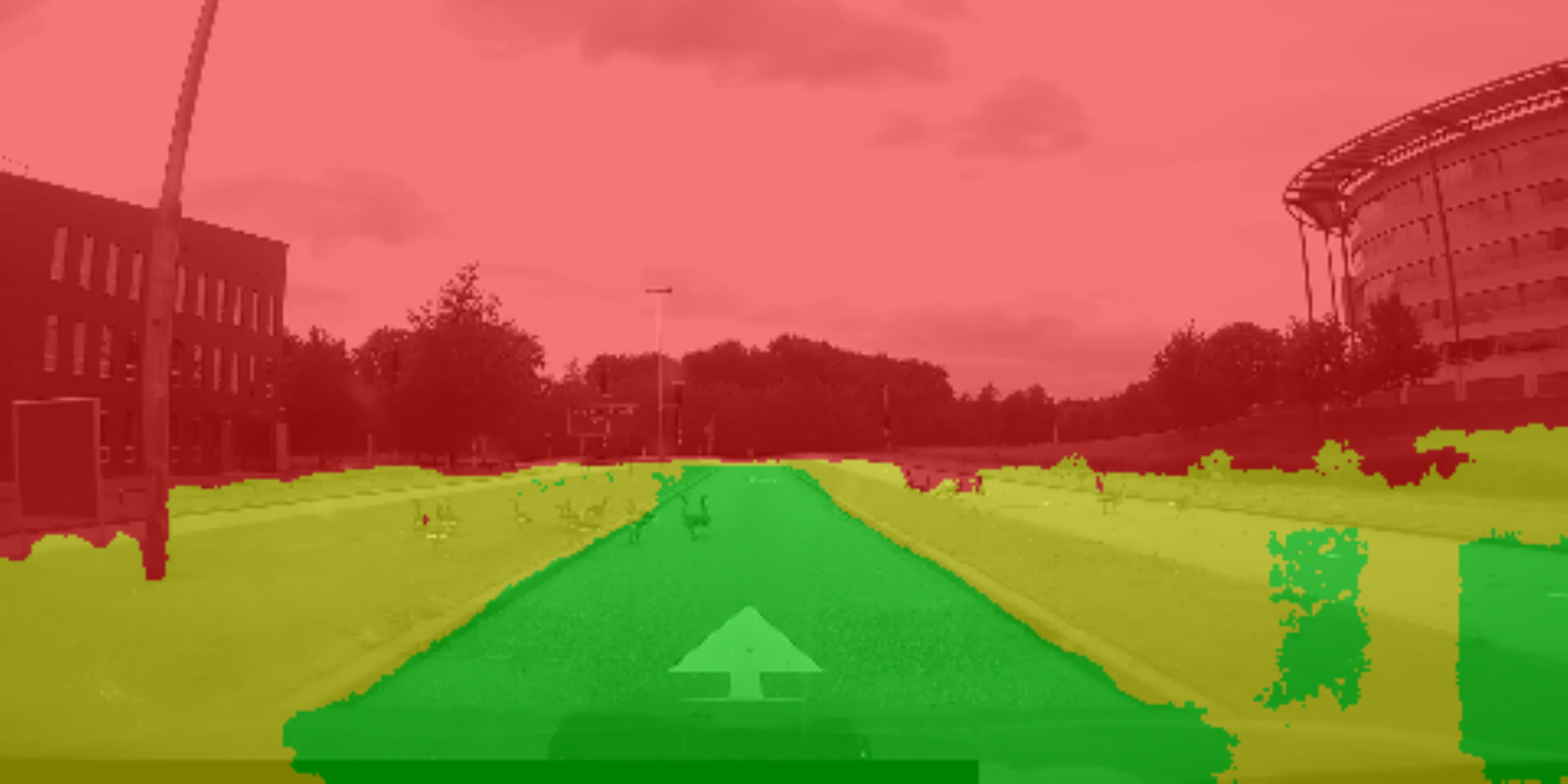}
    \includegraphics[width=0.246\linewidth]{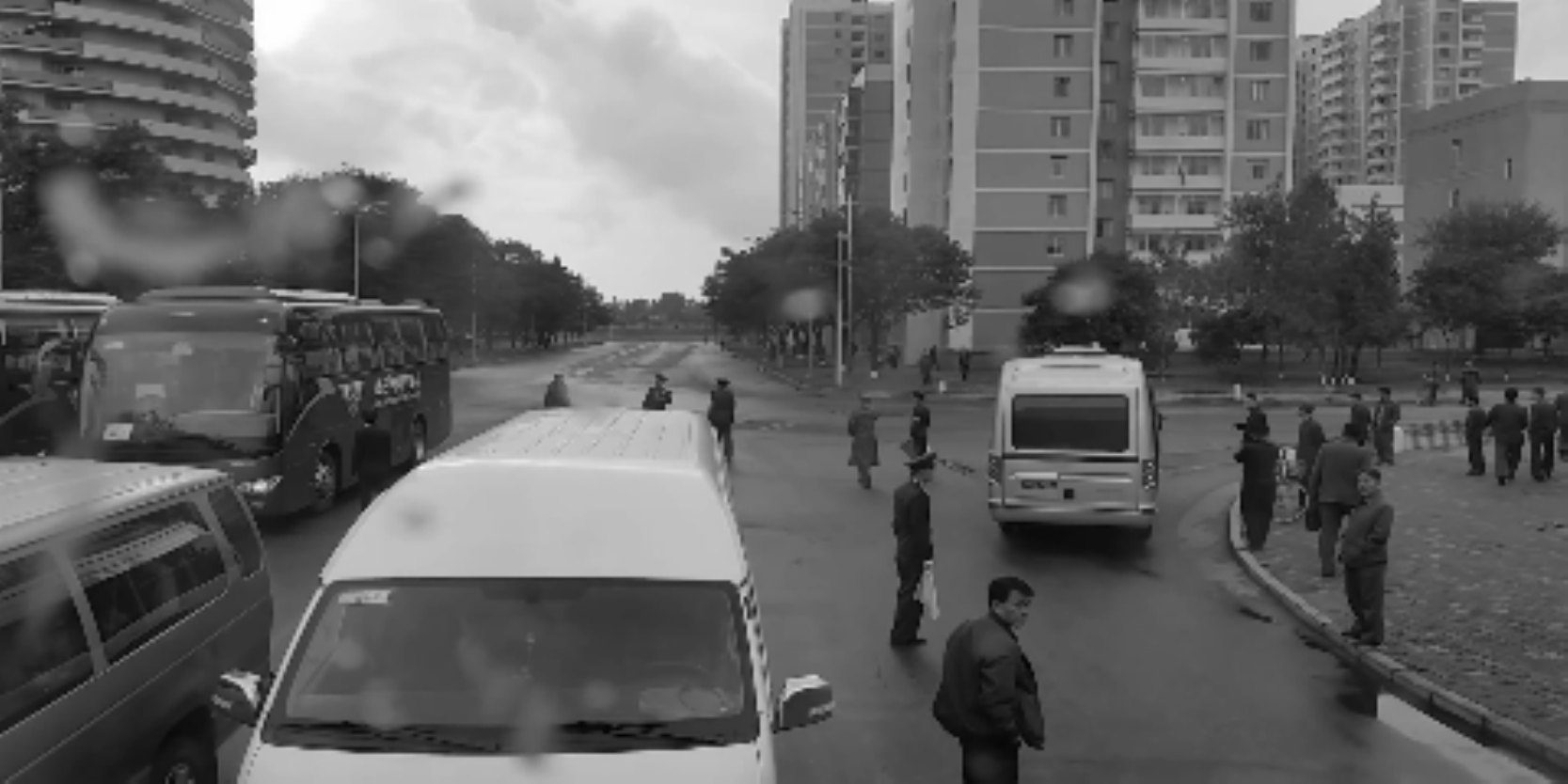}\includegraphics[width=0.246\linewidth]{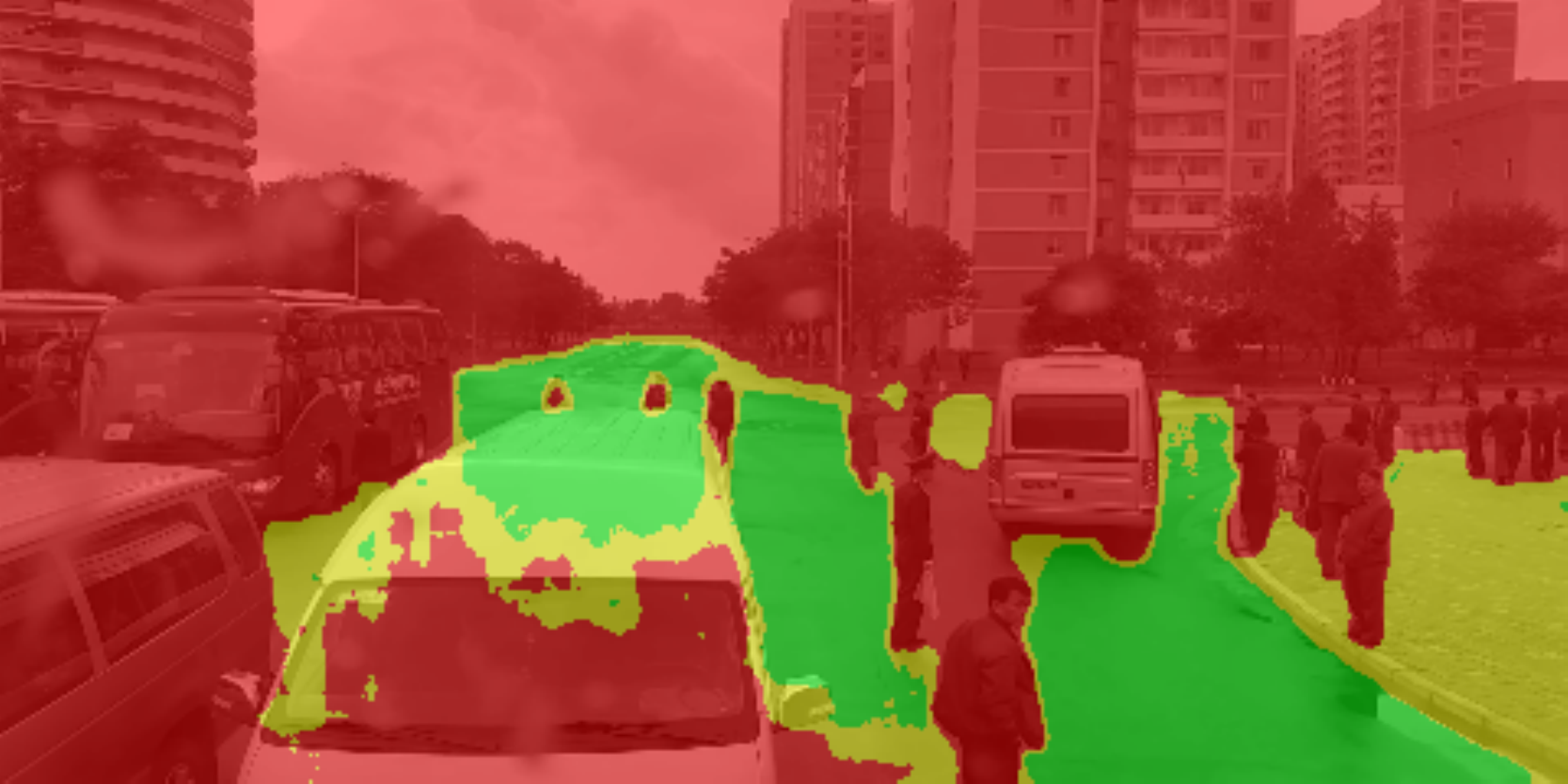}
    \small{\textit{Overlooking important obstacles}}
    
    \caption{Examples of unacceptable predictions on WildDash~\cite{dataset-wilddash-2018} images.}
    \label{fig:bad}
\end{figure}

\textbf{Failure cases:} As indicated by its performance on the Kitti Object \& Tracking datasets (Table~\ref{tab:oxford-kitti-detection}), our model reliably detects common obstacles in ideal road scenes. However, looking at predictions on the challenging WildDash benchmark, we note that the model inherits the limitations of RGB vision, with poor results in extreme weather or illumination conditions. In the first row of Figure~\ref{fig:bad}, the images are too dark and foggy (left) or rainy/snowy (right) to estimate driveability, especially through a windshield with the car's dashboard blocking a large portion of the image. In addition, the bottom row examples of Figure~\ref{fig:bad} suggests that distinguishing flat textures with obstacles can be tricky in case of small, unusual objects (eg. ducks in the bottom left), or structures aligning with the road configuration (bottom right). We expect that the incorporation of depth cues for driveability estimation would help disambiguate between road irregularities such as manholes, shadows, lane markings (all of which are considered driveable \lilbox{green}) from actual hazards on the robot's path.

\begin{figure}[!htb]
    \centering
    \includegraphics[width=\linewidth]{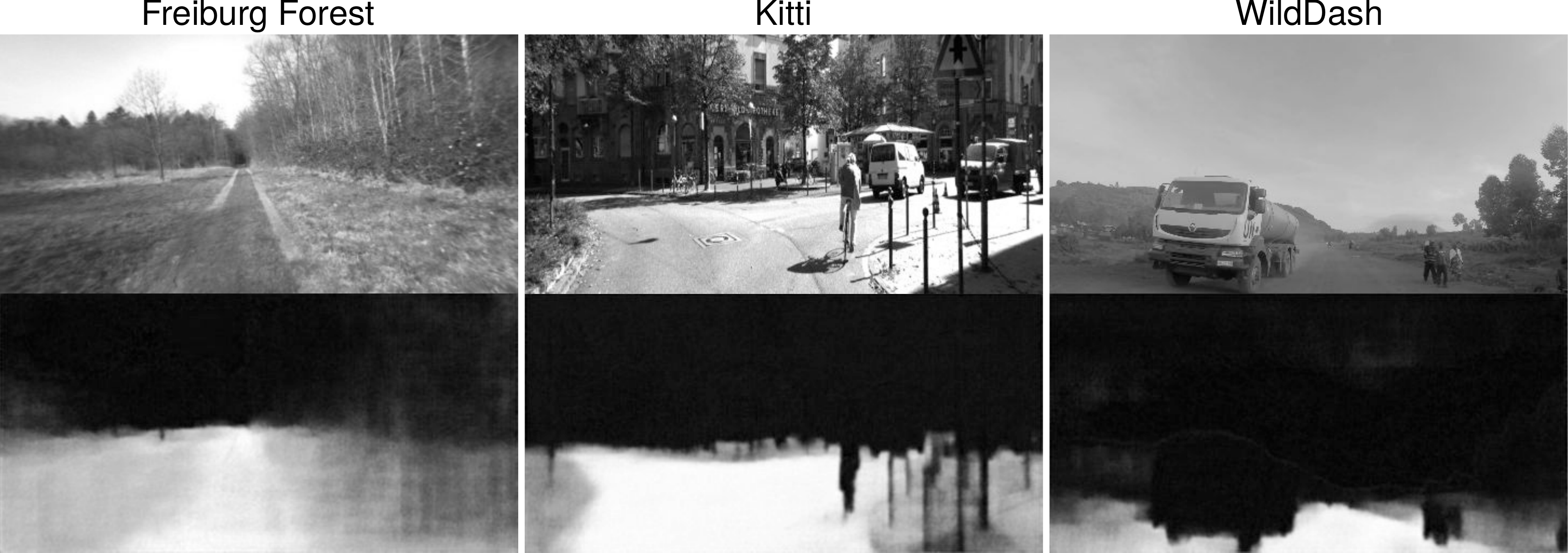}
    \caption{Probabilistic driveability estimation by the Cross-domain SORD+LW model on out-of-dataset samples.}
    \label{fig:proba-maps}
\end{figure}

\textbf{Probabilistic affordance maps for planning:} In our evaluation, we have taken predictions as the argmax of the output layer, resulting in crisp 3-level segmentation. Instead, since our model predicts ordered \textit{ranks}, predictions can also be taken as the expected value $\sum_{\forall r_i \in R} r_i \hat{y}_i$ - resulting in probabilistic affordance maps as shown in Figure~\ref{fig:proba-maps}, with smooth transitions between driveability levels.

\section{Discussion and Future Work}

\textbf{Driveability estimation:}
By defining a simple ground truth mapping between object classes and driveability, we bridge the semantic gap between datasets to allow joint cross-domain training while bypassing the need for manual labelling. However, while this mapping can easily be adapted to the task at hand and robot capabilities, it remains blind to contextual information: the sidewalk may be the \textit{preferable} path for a ``pedestrian'' robot, but only a \textit{possible} last resort for an autonomous vehicle driving on the road; a dirt path may be \textit{preferable} to drive on in a forest, but not a route of choice in a city. Incorporating such scene- and application-dependent context during learning is an important direction for further research. Future work will also investigate how driveability can be learned from multi-modal image data to improve scene understanding in poor visibility.

\textbf{Towards robot navigation:} To investigate the practical implications of our approach for open-world robotic navigation, future work will incorporate our probabilistic driveability maps (Figure~\ref{fig:proba-maps}) into a severity-aware planning module, which aims to maximize driveability along sampled trajectories. To this end, our pixel-wise affordances could be projected into 3D space using depth and odometry data, and used as a cost for graph-based path planning - as demonstrated in \cite{legged-terrain-classification-2019} and \cite{vizdoom-affordance-maps-2020}. For safe navigation in urban environments, our method should also be complemented with recognition of traffic cues~\cite{cv-autonomous-driving-2020}. Extending our 3-level definition to distinguish between static background and moving obstacles may also be beneficial.

\section{Conclusion}

We have presented a simple yet effective method for learning pixel-wise driveability across outdoor scenes for open-world robotic navigation. Compared to existing approaches which treat all pixels and mistakes equally and are constrained to a specific domain, our severity-aware affordance learning framework allows cross-dataset training and tailors the label and loss formulation to navigation, with quantitative and qualitative improvements in segmentation of unseen environments.

\ifCLASSOPTIONcaptionsoff
  \newpage
\fi



%
\bibliographystyle{IEEEtran}
\bibliography{IEEEabrv,refs}

%








\end{document}